\documentclass{article} % For LaTeX2e
\usepackage{iclr2arxiv,times}

\usepackage{amsmath,amsfonts,bm}

\def\Figref#1{Figure~\ref{#1}}

\def\eqref#1{equation~\ref{#1}}
\def\Eqref#1{Equation~\ref{#1}}
\def\Algref#1{Algorithm~\ref{#1}}

\def\1{\bm{1}}

\DeclareMathAlphabet{\mathsfit}{\encodingdefault}{\sfdefault}{m}{sl}
\SetMathAlphabet{\mathsfit}{bold}{\encodingdefault}{\sfdefault}{bx}{n}

\usepackage{hyperref}
\usepackage{url}
\usepackage{graphicx}
\usepackage{booktabs}
\usepackage{multirow}
\usepackage{wrapfig}
\usepackage{enumitem}

\usepackage{algorithm}
\usepackage{algpseudocode}  % 加的
\usepackage{caption}        % 加的
\usepackage{multirow}       % 加的
\usepackage{siunitx} % for number alignment 加的
\usepackage{booktabs} % for top/mid/bottomrule 加的
\usepackage[table]{xcolor} % for row colors 加的
\usepackage{amssymb}%加的

\title{Supervised Fine-Tuning or Contrastive Learning? Towards Better Multimodal LLM Reranking}

\author{
Ziqi Dai$^{1}$\thanks{Equal contribution}, Xin Zhang$^{1,2}$\footnotemark[1], Mingxin Li, Yanzhao Zhang, Dingkun Long \\
Pengjun Xie, Meishan Zhang$^1$, Wenjie Li$^2$, Min Zhang$^1$ \\ \small{$^1$Harbin Institute of Technology, Shenzhen \quad
$^2$The Hong Kong Polytechnic University}
 \\
\texttt{\{ziqi.dai, zhangxin2023\}@stu.hit.edu.cn} \\
Release at \url{https://github.com/vec-ai/lychee-rerank-mm} \\
}

\newcommand{\ie}{\emph{i.e.,}~}
\newcommand{\eg}{\emph{e.g.,}~}
\newcommand{\etc}{\emph{etc.}~}

\newcommand{\resp}{\emph{resp.}~}

\begin{document}

\maketitle

\begin{abstract}
In information retrieval, training reranking models mainly focuses on two types of objectives: metric learning (\eg contrastive loss to increase the predicted scores on relevant query-document pairs) and classification (binary label prediction of relevance vs. irrelevance).
For BERT-style encoders, various studies have shown that contrastive learning (CL) can be more effective than discriminative (classification) learning.
However, for large language models (LLMs), classification via supervised fine-tuning (SFT), which predicts ``yes'' (\resp ``no'') token for relevant (\resp irrelevant) pairs, appears more promising as it aligns well with the generative nature of LLMs.
This divergence raises a central question: \emph{which objective is intrinsically better suited to LLM-based reranking, and what mechanism underlies the difference?}
In this work, we conduct a comprehensive comparison and analysis between CL and SFT for reranking, taking the universal multimodal retrieval (UMR) as the experimental playground.
We first decompose the objectives into two components:
\emph{weight}, which controls the magnitude of those updates,
and \emph{direction}, which guides the model updates,
then present a unified framework for understanding their interactions.
Through probing experiments, we find that SFT provides a substantially stronger weighting scheme than CL, whereas the preferred scoring direction shows no clear winner.
Taken together, these results point to a consistent advantage of SFT over CL for LLM reranking.
To further validate our findings, we conduct large-scale training with SFT and present new state-of-the-art rerankers on the MRB benchmark.
We also provide ablations on SFT settings and expect our findings to benefit future research and applications in this area.
\end{abstract}

\section{Introduction}

Reranking is a crucial step in the retrieval pipeline \citep{lin2022pretrained}, aiming to refine the initial results obtained from the previous search stage by reordering them based on their relevance to a given query.
In recent years, the integration of Large Language Models (LLMs) into reranking techniques has shown promising results in text retrieval \citep{ma2024finetuning} and has gradually become the standard approach \citep{sharifymoghaddam2025rankllm}.
When extending to the multimodal setting \citep{liu2023universal,cong2025uniir}, multimodal LLMs (MLLMs) also become the promising backbone choice \citep{lin2025mmembed,Zhang_2025_CVPR} as their strong multimodal understanding capabilities.

\begin{figure}
\centering
\includegraphics[width=0.9\linewidth]{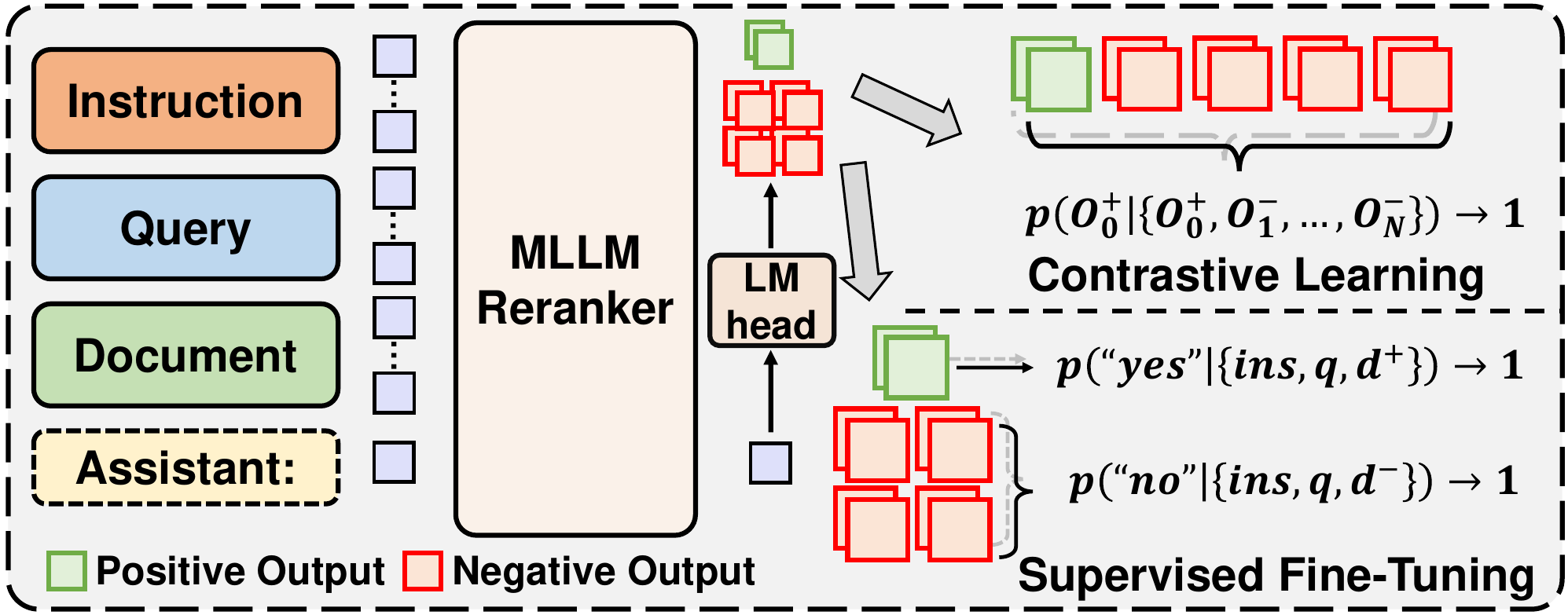}
\vspace{-0.6em}
\caption{
Comparison of Supervised Fine-Tuning (SFT) and Contrastive Learning (CL) for the MLLM reranker.
}
\label{fig:rerank_cl_vs_sft}
\vspace{-1.2em}
\end{figure}

Current widely used rerankers are typically in the point-wise setting \citep{lin2022pretrained}, which independently scores each query-candidate pair and ranks the candidates.
The simple architecture of point-wise rerankers makes them easy and efficient to applicate in real-world scenarios, and there emerges various open-source state-of-the-art (SOTA) models \citep{chen-etal-2024-m3,zhang-etal-2024-mgte}, particularly LLM-based ones \citep{sharifymoghaddam2025rankllm,zhang2025qwen3e}.
To train such rerankers\footnote{
Throughout this work, reranking primarily refers to point-wise reranking setting.
}, one straightforward approach follows the pre-LLM practice of contrastive learning (CL) \citep{nogueira2019multi,zhang-etal-2024-mgte}, computing InfoNCE loss \citep{oord2018representation} one predicted relevance scores.
Another approach is to directly perform supervised fine-tuning (SFT) \citep{nogueira-etal-2020-document,zhang2025qwen3e}, which optimizes the model to predict the next token (``true/yes'' for relevant, ``false/no'' for irrelevant) and takes the ``true/yes'' token probability as the relevance score.
The illustration of them is shown in \Figref{fig:rerank_cl_vs_sft}.
Before the emergence of LLMs, contrastive learning was the dominant approach for leveraging BERT-style encoders due to its strong performance \citep{nogueira2019multi,zhang-etal-2024-mgte}.
However, SFT are now widely applied to LLMs \citep{nogueira-etal-2020-document,zhang2025qwen3e} and appears to deliver competitive results.
This raises a natural research question: \emph{which objective is intrinsically better for LLM reranking, and why?}

Meanwhile, research on multimodal reranking remains largely restricted to single datasets or narrowly defined tasks \citep{xu2025mm}, limiting the generalizability of existing approaches. Building on recent advances in universal multimodal retrieval \citep{Zhang_2025_CVPR}, our objective is to develop a universal multimodal reranking model that can consistently adapt across diverse modalities.

In this work, we aim to explore the question by providing a theoretical analysis and empirical comparison of the two approaches on the universal multimodal retrieval task as testbed.
We first design the \underline{G}eneral \underline{M}ultimodal \underline{R}eranker (GMR, \S\ref{sec:method:gmr}), and then analyze the two training approaches and decompose their loss functions (\S\ref{sec:method:loss_decompose}) into \emph{weight} and \emph{direction}.
Based on this, we implement a unified framework for CL and SFT losses and conduct experiments to compare and analysis them (\S\ref{sec:analysis}).
To make comprehensive evaluations of multimodal reranking, we compile a new unified benchmark  called \emph{MRB} (multimodal reranking benchmark, \S \ref{sec:exp}).

Through analysis and comparison, we find that SFT consistently outperforms CL for LLM-based rerankers, and:
(1) The weight component, rather than the direction, accounts for the most performance gap;
(2) A larger weight improves robustness to numerical errors in training, where SFT intrinsically assigns larger weights than CL;
(3) The function of weight is a input-specific guidance: down-weight already-well-learned input pairs and up-weight hard or under-fit pairs;
(4) The native SFT direction is almost optimal, whereas CL can be further improved by tuning its direction matrix.
To further validate the potential of SFT, we train two reranking models (\ie GMR-3B and GMR-7B), which set new state-of-the-art results on MRB.
We will release code, data and models to facilitate future research in this area.
Our contributions are: 
\begin{itemize}[itemsep=0pt,topsep=0pt,leftmargin=2em]
\item We provide a unified analysis of SFT and CL for LLM-based reranking, showing that SFT intrinsically outperforms CL. By decomposing the loss into \emph{weight} and \emph{direction} components, we reveal that SFT's weight term delivers stronger optimization signals.
\item We introduce the MRB benchmark, comprising 40 datasets across single-, cross-, and fused-modal retrieval, offering a comprehensive evaluation for universal multimodal reranking.
\item We develop GMR models, instruction-aware multimodal LLM rerankers trained on 1.5M diverse pairs. GMR-3B and GMR-7B achieve state-of-the-art results on MRB, highlighting the effectiveness of SFT and providing strong backbones for future research.
\end{itemize}

\section{Related Work}
\paragraph{Reranking with Large Language Model.}
Reranking improves retrieval output quality by jointly modeling the query and retrieved candidates and reorder the candidates \citep{lin2022pretrained}.
In recent years, reranking is dominated by methods based on pretrained language models \citep{nogueira2019multi,nogueira-etal-2020-document}, with LLM-based approaches becoming particularly prominent in the latest advancements \citep{ma2024finetuning,zhuang2024asetwise,sharifymoghaddam2025rankllm}.
Compared to the widely studied list-wise reranking \citep{ren2025self,liu2025leveraging}, in this work, we focus on the more straightforward and widely used \emph{point-wise} approach \citep{zhang-etal-2024-mgte,guo2025mcranker}, which scores each query-candidate pair independently and ranks the candidates.

Training point-wise rerankers has traditionally relied on contrastive learning (CL) \citep{nogueira2019multi,zhang-etal-2024-mgte}, which is also a verified choice for LLM-based models \citep{ma2024finetuning}.
However, for such generative language models, a supervised fine-tuning (SFT) approach \citep{nogueira-etal-2020-document} seems to be more aligned with the model nature, as it directly optimizes the model to predict the next token (``true/yes'' for relevant, ``false/no'' for irrelevant) based on the input query and candidate, rather than relying on a contrastive loss that compares the relevant and irrelevant candidates.
There is no clear consensus on which approach is better yet.
To bridge this significant research gap, we conduct a theoretical analysis with empirical comparison of the two approaches, and demonstrate that SFT outperforms CL in terms of performance.

\paragraph{Multimodal Information Retrieval.}
Multimodal Retrieval aims to retrieve relevant candidates from and based on modalities beyond text \citep{wang2024cross}, which involves various sub-tasks such as image-text retrieval \citep{ijcai2022p759} and composed image retrieval \citep{song2025comprehensive}.
Recent advancements in this field have been shifted to a more generalized view, exploring the universal multimodal retrieval (UMR) \citep{liu2023universal,cong2025uniir,Zhang_2025_CVPR} which compile a wide range of datasets and tasks into a unified benchmark.
Retrievers \citep{lin2025mmembed,Zhang_2025_CVPR} driven by multimodal LLMs have shown significant improvements in understanding and processing multimodal data, enabling more effective retrieval across different modalities.
While the reranking stage is crucial for enhancing the precision of retrieval system, it has been less studied in UMR \citep{lin2025mmembed}.
In this work, we investigate how to build better LLM reranking models, presenting state-of-the-art MLLM-based rerankers for UMR.

\section{Method}\label{sec:method}
In this work we analyze the contrastive learning (CL) and supervised fine-tuning (SFT) approaches to reranking, taking the multimodal retrieval as the experimental playground.
We first introduce our reranking model (\S\ref{sec:method:gmr}), training by CL or SFT (\S\ref{sec:method:train}), and then present our tools for analysis (\S\ref{sec:method:loss_decompose}).

\subsection{Reranker Implementation}\label{sec:method:gmr}

Our general multimodal reranker (namely GMR) follows the conventional design of LLM-based point-wise reranking models.
We employ a strong MLLM as the backbone, which could process diverse input modalities, encompassing images, text, and multimodal combinations.

\paragraph{Instruction-Aware Reranking.}
Given query $q$ and document $d$, we set an instruction $ins$ to describe detailed task objectives, which has proven highly effective in MLLM-based multimodal retrieval \citep{lin2025mmembed,Zhang_2025_CVPR}.
For example, in the Visual Document Retrieval task \citep{ma-etal-2024-unifying,faysse2025colpali}, we use an instruction ``\textit{Find a screenshot that relevant to the user's question.}'' to guide the model to better evaluate the relevance between  query and visual document.
We list all instructions of our model in Appendix \ref{app:model_settings:GMR}.
The inputs are in the form of $(ins, q, d)$ and formatted into the template shown in \Figref{fig:input_prompt} before being fed into the MLLM backbone.

\paragraph{Relevance Score Computation.}
In the \textbf{SFT} setting,
given the task instruction $ins$, query $q$ and document $d$, our reranker assesses the probability of the next token being either ``yes'' or ``no'' to be the relevance score $\sigma$.
This process could be formally expressed as:
\begin{equation}
\sigma(ins,q,d)=\frac{e^{P(\text{``yes''} |\{ins, q, d\})}}{e^{P(\text{``yes''} |\{ins, q, d\})}+e^{P(\text{``no''} |\{ins, q, d\})}} ,
\label{eq:sft_score}
\end{equation}
where $P(\text{``yes''} | \{ins,q,d\})$ and $P(\text{``no''} | \{ins,q,d\})$ represent the probabilities of the next token being ``yes'' or ``no'', respectively, given the document and query as context.
With such relevance scores, we could rerank all retrieved candidates more precisely.
This method is more aligned to the generative nature of MLLM and thus allows us to leverage its powerful understanding ability while providing a effective scoring mechanism for reranking purposes.
In the \textbf{CL} setting, the relevance score is the ``yes'' probability only:
% , \ie $s(i,q,d)=P(\text{``yes''} |\{i, q, d\})$.
\begin{equation}
\sigma(ins,q,d)=P(\text{``yes''} |\{ins, q, d\}) .
\label{eq:cl_score}
\end{equation}

\subsection{Reranker Training}\label{sec:method:train}
In reranking, each data example contains one query $q$, one relevant document (positive) $d^+_0$, and several irrelevant documents (negatives, the selection is described in Appendix \ref{app:exp:neg}) $\{d^-_1,d^-_2,\dots,d^-_N\}$.
As shown in \Figref{fig:rerank_cl_vs_sft}, we explore both CL and SFT based training.

$\bullet$ \textbf{Contrastive Learning:} 
With relevance score $\sigma$ from \Eqref{eq:cl_score}, we compute the InfoNCE loss \citep{oord2018representation} for each example:
\begin{equation}\label{eq:cl_loss}
    \mathcal{L}^\mathrm{CL}=-\log\frac{\exp(\sigma(ins, q, d_0^+))}{\exp(\sigma(ins, q, d_0^+))+\sum_i\exp(\sigma(ins, q, d_i^-))} .
\end{equation}

$\bullet$ \textbf{Supervised Fine-Tuning:}
The objective is predicting correct next token (relevance label) for each input pair, independently.
We reorganize one example into multiple triples (${ins, q, d_i}$), each corresponding to a different $d$.
Then predict the likelihood of ``yes'' and ``no'' for each triplet and compute per-triplet cross-entropy loss with the token of ground-truth label $l$:
\begin{equation}
    \mathcal{L}^\mathrm{SFT}_i = -\log(p(l|P(\{\text{``yes'',``no''}\}|\{ins,q,d_i\}))) ,
\label{eq:loss_function_sft}
\end{equation}
where $P(\{\text{``yes'',``no''}\}|\{ins,q,d\})$ denotes the likelihood of ``yes'' and ``no''. The relevance label $l$ is ``yes'' for positive documents and ``no'' for negatives.
This loss encourages the model to assign higher probabilities to correct tokens, thereby improving the ranking performance.

\subsection{Loss Function Decomposition}
\label{sec:method:loss_decompose}
We analyze two reranking loss functions by decomposing them into two key components: \emph{weight} and \emph{direction}.
With this decomposition, we perform probing experiments in \S\ref{sec:analysis}.

\paragraph{Basic Notation.}
We denote the SFT-style data instance with positive (\resp negative) doc as $o^+_0 = \{ins,q,d^+_0\}$ (\resp $o^-_j = \{ins,q,d^-_i\},i=1,2,\dots,N$).
The reranker is conceptualized as two components:
a mapping function $f(\cdot|\theta)$ (parameterized by $\theta$) that converts $o_i$ to the feature representation $\mathbf{h}_i = f(o_i|\theta)$,
and a transformation $\mathcal{M}^y$ that maps $\mathbf{h}_i$ into ``yes'' token likelihood score ${s}^y (h_i) = \mathbf{h}_i \cdot \mathcal{M}^y$.
And the ``no'' token score ${s}^n$ could be computed similarly by $\mathcal{M}^n$.

\paragraph{Unified View.}
From \S\ref{sec:method:train}, the SFT loss is calculated separately for each positive or negative doc of an example, while the CL loss is computed in an integrated manner across all positive and negative docs of the same example.
To enable a fair comparison, we adopt the total loss $\mathcal{L}(\{o_i\}_{i=0}^N,\theta)$ over an entire example (with one positive and $N$ negatives) as the unit of analysis.
So we have the gradient
\begin{equation}
\frac{\partial \mathcal{L}}{\partial\theta}=\frac{\partial\mathcal{L}}{\partial \mathbf{h}_0^+}\frac{\partial \mathbf{h}_0^+}{\partial \theta}+\sum_{i}\frac{\partial\mathcal{L}}{\partial \mathbf{h}^-_i}\frac{\partial \mathbf{h}^-_i}{\partial \theta} ,
\end{equation}
where $\mathbf{h}_0^+$ is the feature of positive doc and $\mathbf{h}^-_i$ is that of $i$-th negative.

To understand the influence of positive and negatives on the model, we calculate the partial derivative of the loss function with respect to the hidden state.
For CL, we only use ``yes'' token, and by substituting the specific loss (\Eqref{eq:cl_loss}) into the gradient, we obtain the partial derivatives:
\begin{equation}
-\frac{\partial\mathcal{L}^\mathrm{CL}}{\partial \mathbf{h}_0^+}
=\frac{\sum_j\exp(s^y(\mathbf{h}^-_i))}{\exp(s^y(\mathbf{h}_0^+))+\sum_i\exp(s^y(\mathbf{h}^-_i))}\mathcal{M}^y ,
\label{eq:cl_pos}
\end{equation}
\begin{equation}
-\frac{\partial\mathcal{L}^\mathrm{CL}}{\partial \mathbf{h}^-_i}
=-\frac{\exp(s^y(\mathbf{h}^-_i))}{\exp(s^y(h_0^+))+\sum_i\exp(s^y(\mathbf{h}^-_i))}\mathcal{M}^y .
\label{eq:cl_neg}
\end{equation}

In SFT, we first merge the \Eqref{eq:loss_function_sft} of multiple pairs in one example into the total loss
\begin{equation}
\mathcal{L}^\mathrm{SFT} = -\log \frac{\exp(s^y(\mathbf{h}_0^+))}{\exp(s^y(\mathbf{h}_0^+))+\exp(s^n(\mathbf{h}_0^+))} \nonumber -\sum_i\log \frac{\exp(s^n(\mathbf{h}^-_i))}{\exp(s^y(\mathbf{h}^-_i))+\exp(s^n(\mathbf{h}^-_i))} .
\end{equation}
Then we have partial derivatives
\begin{equation}\label{eq:sft_pos}
-\frac{\partial\mathcal{L}^\mathrm{SFT}}{\partial \mathbf{h}_0^+}
=\frac{\exp(s^n(\mathbf{h}_0^+))}{\exp(s^y(\mathbf{h}_0^+))+\exp(s^n(\mathbf{h}_0^+))}
(\mathcal{M}_\mathrm{y}-\mathcal{M}_\mathrm{n}) ,
\end{equation}
\begin{equation}
-\frac{\partial\mathcal{L}^\mathrm{SFT}}{\partial \mathbf{h}^-_i}
=-\frac{\exp(s^y(\mathbf{h}^-_i))}{\exp(s^y(\mathbf{h}^-_i))+\exp(s^n(\mathbf{h}^-_i))}
(\mathcal{M}_y-\mathcal{M}_n) .
\label{eq:sft_neg}
\end{equation}
The complete derivation of the above process is provided in the Appendix \ref{ap:loss_function_decomposition}.

\begin{wrapfigure}{r}{0.45\textwidth}
\vspace{-2\baselineskip}  
\begin{minipage}[t]{0.45\textwidth}
\begin{algorithm}[H]  
\caption{Unified Reranking Loss}
\label{alg:wd-loss}
\begin{algorithmic}[1]              
  \Require inputs $\mathcal{O}\gets\{o^{+}_{0},\,\dots,o^{-}_{n}\}$ %,
        %   modes \texttt{weight}, \texttt{direction} $\in\{\text{sft},\text{cl}\}$
  \Ensure   loss value $\mathcal{L}$
  \State $\mathcal{M} \gets \texttt{lm\_head}(\text{``yes'',``no''})$
  \State $logits \gets \mathcal{M} \cdot f(\mathcal{O}|\theta)$
  % ------------------------------------------------------
  \Statex\textit{//--- weight branch ----------------------------}
  \If{\texttt{weight}$=\,$``sft''}
      \State $s\gets\mathrm{Softmax}(logits)[0].detach()$   
      \State $W^{+}\gets W^{+}_{\text{sft}}\!\gets 1-s[0]$
      \State $W^{-}\gets W^{-}_{\text{sft}}\!\gets s[1{:}]$
  \Else\Comment{\texttt{weight}$=\,$``cl''}
      \State $s\gets\mathrm{Softmax}(logits[0]).detach()$
      \State $W^{+}\gets W^{+}_{\text{cl}}\gets 1-s[0]$
      \State $W^{-}\gets W^{-}_{\text{cl}}\gets s[1{:}]$
  \EndIf
  % ------------------------------------------------------
  \Statex\textit{//--- direction branch ----------------------}
  \State $M_{\mathrm{y}}\gets logits[:,0]$;\quad
         $M_{\mathrm{n}}\gets logits[:,1]$
  \If{\texttt{direction}$=\,$``sft''}
      \State $D^{+}\gets D^{+}_{\text{sft}}\gets M_{\mathrm{n}}[0]-M_{\mathrm{y}}[0]$
      \State $D^{-}\gets D^{-}_{\text{sft}}\gets M_{\mathrm{y}}[1{:}]-M_{\mathrm{n}}[1{:}]$
  \Else\Comment{\texttt{direction}$=\,$``cl''}
      \State $D^{+}\gets D^{+}_{\text{cl}}\gets -\,M_{\mathrm{y}}[0]$
      \State $D^{-}\gets D^{-}_{\text{cl}}\gets M_{\mathrm{y}}[1{:}]$
  \EndIf
  % ------------------------------------------------------
  \Statex\textit{//----------------------------------------------------}
  \State $\mathcal{L}\gets
         \operatorname{mean}\!\bigl(W^{+}D^{+}+\sum\nolimits_{i}W^{-}_{i}D^{-}_{i}\bigr)$
  \State \Return $\mathcal{L}$
\end{algorithmic}
\end{algorithm}
% \vspace{-2em}  % 调整垂直间距
\end{minipage}
\vspace{-10em}  % 调整垂直间距
\end{wrapfigure}

\paragraph{Loss Decomposition.}
As above gradients looks similar, we can break them down into two parts: \emph{weight}
 and \emph{direction}.
They reflect the differences between CL and SFT.

$\bullet$ \emph{Weight} $W$ is a scalar that controls the magnitude of the updates. From \Eqref{eq:cl_pos} - \ref{eq:sft_neg}, 
we obtain the weights as shown below:
\begin{align}
W_\mathrm{CL}^+ &= \frac{\sum_i\exp(s^y(\mathbf{h}^-_i))}{\exp(s^y(\mathbf{h}_0^+))+\sum_i\exp(s^y(\mathbf{h}^-_i))} , \label{eq:cl_pos_weight} \\
W_\mathrm{CL}^- &= \frac{\exp(s^y(\mathbf{h}^-_i))}{\exp(s^y(\mathbf{h}_0^+))+\sum_i\exp(s^y(\mathbf{h}^-_i))} ,
\label{eq:cl_neg_weight} \\
W_\mathrm{SFT}^+ &= \frac{\exp(s^n(\mathbf{h}_0^+))}{\exp(s^y(\mathbf{h}_0^+))+\exp(s^n(\mathbf{h}_0^+))} , 
\label{eq:sft_pos_weight} \\
W_\mathrm{SFT}^- &= \frac{\exp(s^y(\mathbf{h}^-_i))}{\exp(s^y(\mathbf{h}^-_i))+\exp(s^n(\mathbf{h}^-_i))} .
\label{eq:sft_neg_weight} 
\end{align}
Compared with CL, $W_{SFT}$ only focus on the single document, without the interactions with all negatives of the same query like CL.

$\bullet$ \emph{Direction} $D$ is a vector that controls the direction of model updates. From \Eqref{eq:cl_pos} and \ref{eq:sft_pos}, the direction from the positive $d^+$ for CL is $D_\mathrm{CL}^+ = \mathcal{M}_\mathrm{y}$, and that of SFT is $D_\mathrm{SFT}^+ = \mathcal{M}_\mathrm{y}-\mathcal{M}_\mathrm{n}$.
While from \Eqref{eq:cl_neg} and \ref{eq:sft_neg}, direction from negatives are $D_\mathrm{CL}^-  = -\mathcal{M}_\mathrm{y}$ and $D_\mathrm{SFT}^- = -(\mathcal{M}_\mathrm{y}-\mathcal{M}_\mathrm{n})$.
Apparently, for both CL and SFT, the update directions of positive and negatives are opposite.

In summary, CL and SFT share similar direction components, and we believe that differing initializations\footnote{
$\mathcal{M}_\mathrm{y}$ compared to $\mathcal{M}_\mathrm{y}-\mathcal{M}_\mathrm{n}$.
} are insufficient to account for performance differences.
In contrast, \emph{CL computes the weight using all positive and negative documents within a sample, while SFT assigns weights independently per document}, making this the likely key factor in performance variation.

\paragraph{Unified Framework}
Building on the above decomposition, we propose a unified reranking loss framework (\texttt{URL}), with pseudo-code provided in \Algref{alg:wd-loss}. This framework allows us to independently analyze \emph{weight}
 and \emph{direction}, thereby facilitating a deeper understanding of the differences between the two training paradigms through controlled adjustments during computation. 
We then validate our analysis through probing experiments in the following \S\ref{sec:analysis}.

\section{Analysis}\label{sec:analysis}

In this section, we continue and validate the analysis in \S\ref{sec:method:loss_decompose} through probing experiments.
We choose universal multimodal retrieval as the testbed, compiling a new benchmark (MRB \S\ref{sec:exp:setting}) includes single-modal tasks (text-to-text, image-to-image), cross-modal tasks (\eg text-to-image), as well as fused-modal tasks (either the query or the document could consist of text + image).
\begin{wrapfigure}{r}{0.3\textwidth} 
\vspace{-0.3em}  
\includegraphics[width=0.3\textwidth]{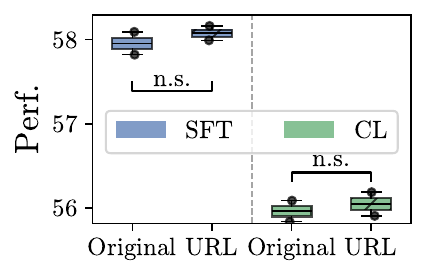}
\vspace{-1.8em}  
\caption{Performance comparison of the original implementations and our \texttt{URL}.}
\label{fig:impl-cmp}
\vspace{-2.5em}  
\end{wrapfigure}
We defer the description of experiment settings and evaluation benchmark to \S\ref{sec:exp:setting}.

\paragraph{General Empirical Comparison.}
We first train both CL and SFT rerankers with the original implementation and our \texttt{URL} framework to (1) find the winner in practice,
and (2) verify that \texttt{URL} faithfully reproduces the original implementation, supporting the subsequent analyses built on \texttt{URL}.
As shown in \Figref{fig:impl-cmp}, under the identical setting, \emph{SFT consistently outperforms CL}.
Meanwhile, \texttt{URL} yields statistically indistinguishable performance to the original.
It thus could be trusted in the following analysis.

\begin{wraptable}{r}{0.34\textwidth}
\vspace{-1em}  % 调整垂直间距
\centering
\setlength{\tabcolsep}{3pt} % 调整列间距
\sisetup{table-format=2.2}  % 小数点对齐
\begin{tabular}{l S S S}
\toprule
 & {$\mathit{D}_{\text{SFT}}$} & {$\mathit{D}_{\text{CL}}$} & {\textbf{$\Delta_\mathit{D}$}} \\
\midrule
$\mathit{W}_{\text{SFT}}$ & 58.09 & 57.88 & {\textcolor{red}{$\blacktriangledown$}~{0.21}} \\
$\mathit{W}_{\text{CL}}$  & 56.99 & 56.40 & {\textcolor{red}{$\blacktriangledown$}~{0.59}} \\
\rowcolor{gray!15}
$\Delta_\mathit{W}$       & {\textcolor{red}{$\blacktriangledown$}~\textbf{1.10}} & {\textcolor{red}{$\blacktriangledown$}~\textbf{1.48}} &  \\
\bottomrule
\end{tabular}
\caption{
MRB results of all loss components combinations, where the weight $W$ delivers the dominant influence on performance.
}
\label{tab:components}
% \end{table}
\vspace{-2em}
\end{wraptable}

\paragraph{Weight $W$ Dominates Performance.}
To investigate why SFT outperforms CL, we first dissect the contribution of weight and direction.
In Table \ref{tab:components}, we train the model with all combinations by \texttt{URL}.
We observe that the improvements from weight (\ie $\Delta_W$) is more significant than that of direction ($\Delta_D$).
This suggests that the weight $W$ is the dominant factor in the performance gap between SFT and CL, guiding us to focus on the weight in the following section.
However, the direction also contributes to the gap, which is investigated in \S\ref{sec:analysis:direction}.

\subsection{Function of Weight}
\label{sec:analysis:weight}
To figure out why $W_\mathrm{CL}$ is less effective that $W_\mathrm{SFT}$ and what is the function of $W$, we start from the observation of \citep{chen2021simplerfasterstrongerbreaking}.
In small-batch CL training with InfoNCE, \emph{gradients would shrink to very small scale, close to random precision errors, 
and thus cease to provide effective learning guidance}.
We suppose this is more salient in reranking where the small batch size is common\footnote{
Consider a batch of instances, $\{O_1,\dots,O_j\}$, is forward simultaneously during training with $k$ negatives per sample. While dense retrieval can achieve the negative size of $j\cdot (k+1)$ per instances, reranking models' are limited to $k+1$. Furthermore, the increased number of input tokens at the reranking stage, compared to dense retrieval, imposes additional constraints on memory usage, resulting in a reduction in the value of negative size.
}.
Then we validate their findings by training a CL model with fully half-precision loss computation, which yields degraded performance compared to precision-safe training (refer to Appendix \ref{app:analysis:weight}).

\begin{figure}[h]
\centering
\vspace{-1em}
\includegraphics[width=0.8\linewidth]{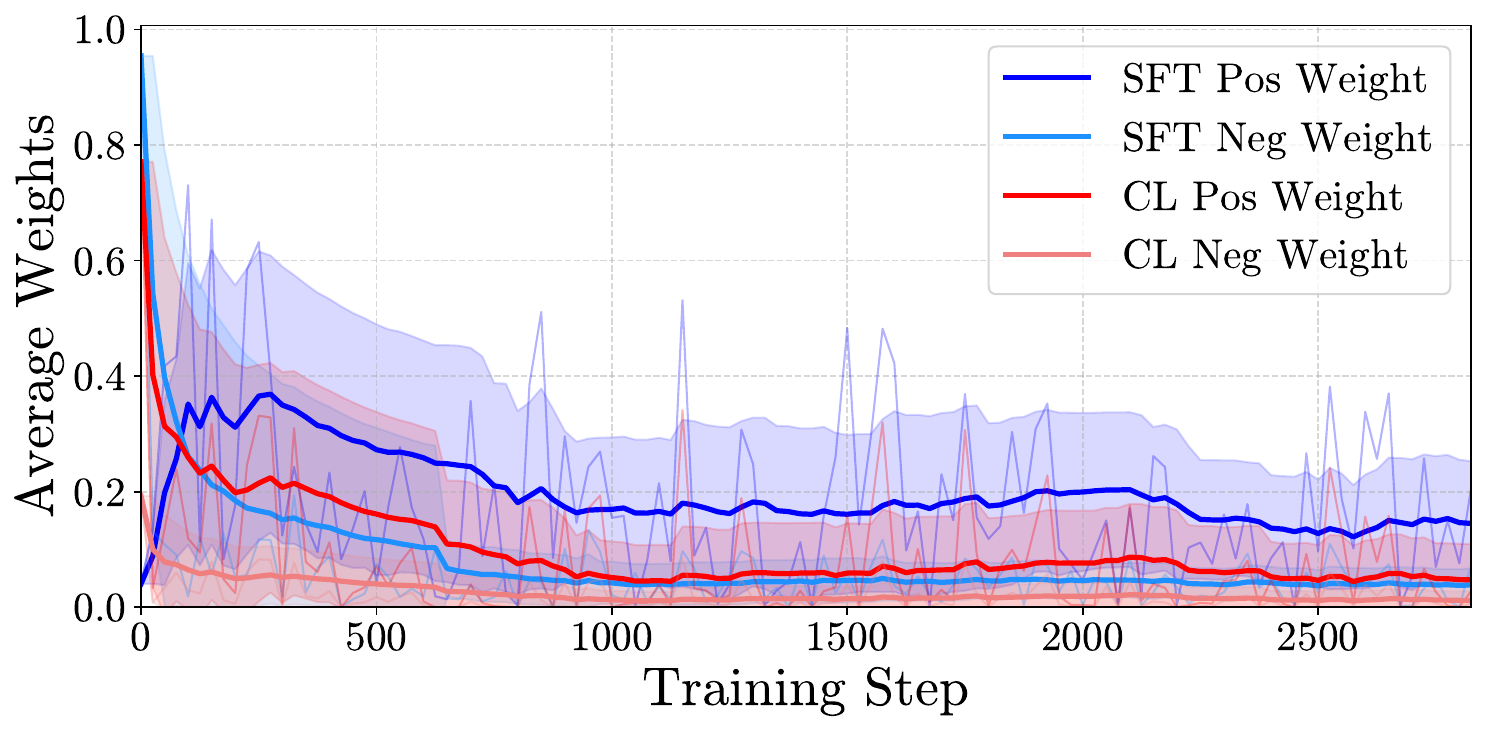}
\vspace{-1em}
\caption{Evolution of positives and negatives average weights during training for SFT and CL. }
\label{fig:weight_vs_step}
\vspace{-1em}
\end{figure}

Back to our framework, $W$ controls the steps of model updates, or say the gradient scale.
According to \cite{chen2021simplerfasterstrongerbreaking}, $W_\mathrm{CL}$ should be small in the training process.
And we expect $W_\mathrm{SFT}$ to be larger than $W_\mathrm{CL}$ to provide better optimization signal as SFT presents better performance.
To verify this, we plot the $W$ of CL and SFT in training in \Figref{fig:weight_vs_step}, where $W_\mathrm{CL}$ indeed show relatively small values.
SFT provides larger (better) $W$ than CL, thereby achieving stronger empirical performance.
\Eqref{eq:cl_pos_weight} to \ref{eq:sft_pos_weight} also shows that $W_\mathrm{SFT}$ is larger than $W_\mathrm{CL}$, since the denominator of $W_\mathrm{CL}$ involves a sum of all negatives while the denominator of $W_\mathrm{SFT}$ only adds up current instance.

Next, we investigate the fine-grained function of $W$.
To create a cleaner analysis setting, we fix the direction in \texttt{URL} as $D_\mathrm{SFT}$ unchanged, as it performs better.
We first set weights of both positive and negatives to the fixed constant $1$ as a baseline ($W_{base}$) following \citep{chen2021simplerfasterstrongerbreaking}:
\begin{align}
    W^+ &= 1, W^-_j=\frac{exp(s(\mathbf{h}^-_j))}{\sum_j exp(s(\mathbf{h}^-))}, \sum_j W^-_j = 1.
\label{function:w_base}
\end{align}
Although the earlier analysis suggests that the larger $W$ is preferable, this value $1$ never appears in \Figref{fig:weight_vs_step}, so we expect this setting to perform poorly.
The experiment in Table \ref{tab:weight_tab1} also align this.
Hence, we suppose that $W$ should be in a reasonable range.
Meanwhile, the failure of constant $W$ indicates that instance-specific adjustment is necessary: \emph{the model should update less on already-mastered instances and more on those it has not yet grasped}.

\begin{wraptable}{r}{0.35\textwidth}
\vspace{-1em}
\centering
\setlength{\tabcolsep}{4pt}
\renewcommand{\arraystretch}{1.1}
\begin{tabular}{c l S c}
\toprule
\textbf{No.} & \textbf{Method} & {\textbf{Avg}} & {$\boldsymbol{\Delta}$} \\
\midrule
1 & $W_{\mathrm{Base}}$       & 49.47 & – \\
2 & + $\tau$ mask             & 56.57 & {\textcolor{green}{$\blacktriangle$}~7.10} \\
3 & + $W_{\mathrm{CL}}$       & 56.23 & {\textcolor{green}{$\blacktriangle$}~6.76} \\
4 & + $W_{\mathrm{SFT}}$      & 58.19 & {\textcolor{green}{$\blacktriangle$}~8.72} \\
\bottomrule
\end{tabular}
\caption{
Evaluation of weight properties. 
$\Delta$ denotes performance gain relative to $W_{\mathrm{Base}}$.
}
\label{tab:weight_tab1}
\vspace{-1em}
\end{wraptable}
We adopt the predicted relevance scores $s$ as a guide and apply a masking rule: if a positive score is high enough, \ie $s(h_0) > 1-\tau$, (or, conversely, a negative score is low enough, $s(h_j) < \tau$), we set $W^+ = 0$ (\resp $W^-_j = 0$) to halt further learning on that instance.
In addition, we further set $W_\mathrm{CL}$ and  $W_\mathrm{SFT}$ to the baseline and conduct training under the same conditions. The results are shown in Table \ref{tab:weight_tab1}, we can see that the simple masking rule can provide strong performance, comparable to CL.
This indicates both CL and SFT follow the above instance-specific weight feature. More details of the experiment can be found in the Appendix \ref{app:analysis:weight}.

\subsection{Searching Better Direction}\label{sec:analysis:direction}

Results in Table \ref{tab:components} indicate that the direction component also affects model performance, but it is not the dominant factor.
Here we conduct additional experiments and try to find a better direction.

\begin{wrapfigure}{r}{0.3\textwidth}  % r表示右侧，0.5\textwidth控制宽度
\vspace{-2em}  % 调整垂直间距
\includegraphics[width=0.3\textwidth]{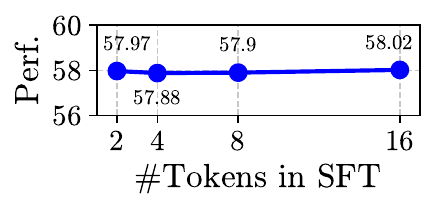}
\vspace{-2em}  % 调整垂直间距
\caption{Results with different token numbers in SFT. The setting with 2 tokens is the standard SFT training.}
\label{fig:sft-token}
\vspace{-1em}  % 调整垂直间距
\end{wrapfigure}

\paragraph{Does adding more tokens improve performance?}
SFT-based training is actually a binary classification on the token labels, where $D_\mathrm{SFT}$ only involves ``yes'' and ``no'' tokens.
One natural question is whether adding more tokens (\eg ``true'', ``false'', ``maybe'', \etc) during training could improve the direction component and model performance?
To investigate this, we randomly select 10,000 training instances and identify the top 16 tokens with the highest logits from the model's output, including ``yes'' and ``no''.
For a comprehensive list of these tokens and details, please refer to the Appendix \ref{app:analysis:token}.
We then train the model using this expanded token set.
\Figref{fig:sft-token} presents the results, which indicate that increasing the number of tokens does not significantly impact model performance.
This result suggests that using only ``yes'' and ``no'' tokens is sufficient for effective SFT.

\begin{wraptable}{r}{0.375\textwidth}
\vspace{-1.2em}
\centering
\setlength{\tabcolsep}{4pt}
\sisetup{table-format=2.2, table-align-text-post=false}

\begin{tabular}{l l S c}
\toprule
\textbf{Weight} & \textbf{Direction} & {\textbf{Perf.}} & {$\boldsymbol{\Delta}$} \\
\midrule
\multirow{2}{*}{$W_{\mathrm{SFT}}$}
  & $D_{\mathrm{SFT}}$   & 58.09 & – \\
  & $D_{\mathrm{Rand.}}$ & 56.75 & {\textcolor{red}{$\blacktriangledown$}~1.34} \\
\midrule
\multirow{2}{*}{$W_{\mathrm{CL}}$}
  & $D_{\mathrm{CL}}$    & 56.40 & – \\
  & $D_{\mathrm{Rand.}}$ & 57.72 & {\textcolor{green}{$\blacktriangle$}~1.32} \\
\bottomrule
\end{tabular}
\caption{
Performance comparison of SFT and CL directions against random initialization $D_{\mathrm{Rand.}}$. 
}
\label{tab:direction_random}
\vspace{-0.8em}
\end{wraptable}

\paragraph{Is it possible to learn a better direction?}
\label{random_direction}
The direction components, in essence, corresponds to the token embeddings of the LLM, which are pre-trained and keeping frozen during training.
Before LLM, CL-based rerankers often learn a score-projection matrix from scratch.
To see whether this still helps, we implement the random-initialized learnable weight $D_\mathrm{Rand.}$ in \texttt{URL}.
Table \ref{tab:direction_random} shows that, for CL models, it does improve performance, yet still trails behind SFT.
For SFT models, however, the strategy hurts performance.
This is in line with the intuition: SFT is trained to predict the “yes/no” tokens, so replacing the weight with a randomly-initialized projection will loss the semantic signal from the pre-trained token embeddings.

\section{Experiments}\label{sec:exp}

\begin{table*}[h]
    \centering 
    \setlength{\tabcolsep}{3pt}
    \resizebox{0.98\linewidth}{!}{
    \begin{tabular}{@{}llccccccccccc@{}}
        \toprule
         Model & Size & \multicolumn{2}{c}{Single-Modal} & \multicolumn{3}{c}{Cross-Modal} & \multicolumn{4}{c}{Fused-Modal} & Avg \\ 
        \cmidrule(lr){3-4} 
        \cmidrule(lr){5-7} 
        \cmidrule(lr){8-11} 
        \cmidrule(lr){12-12} 
        & & T$\rightarrow$T$_{\text{(14)}}$ & I$\rightarrow$I$_{\text{(1)}}$ 
          & T$\rightarrow$I$_{\text{(4)}}$ & T$\rightarrow$VD$_{\text{(5)}}$ & I$\rightarrow$T$_{\text{(5)}}$ 
          & T$\rightarrow$IT$_{\text{(2)}}$ & IT$\rightarrow$T$_{\text{(4)}}$  
          & IT$\rightarrow$I$_{\text{(2)}}$ & IT$\rightarrow$IT$_{\text{(3)}}$ & ALL$_{\text{(40)}}$ \\
        \midrule
        GME-2B & 2.21B & 49.59 & 30.75 & 48.46 & 66.39 & 52.62 & 77.02 & 39.88 & 36.70 & 66.89 & 52.54\\
        \midrule \midrule
        \textit{Qwen3} & 4.02B & \cellcolor{gray!15}\textbf{60.49} & -- & -- & -- & -- & -- & -- & -- & -- & --\\
        \textit{Jina-m0} & 2.21B & 55.36 & 27.50 & \cellcolor{gray!15}\textbf{59.46} & \cellcolor{green!15}\textbf{73.13} & 55.43 & 74.95 & 27.82 & 37.65 & 51.54 & 54.36 \\
        \textit{MonoQwen} & 2.21B & 48.89 & 12.59 & 58.73 & 71.29 & 19.62 & 76.46 & 14.35 & 31.75 & 35.83 & 44.20 \\
        \cmidrule(lr){1-12}
        GMR-3B & 3.75B & 59.22 & \cellcolor{gray!15}\textbf{29.76} & 58.85 & 72.38 & \cellcolor{gray!15}\textbf{63.06} & \cellcolor{gray!15}\textbf{81.96} & \cellcolor{gray!15}\textbf{48.81} & \cellcolor{gray!15}\textbf{43.97} & \cellcolor{gray!15}\textbf{79.08} & \cellcolor{gray!15}\textbf{61.40} \\
        GMR-7B & 8.29B & \cellcolor{green!15}\textbf{61.08} & \cellcolor{green!15}\textbf{32.83} & \cellcolor{green!15}\textbf{61.18} & \cellcolor{gray!15}\textbf{72.94} & \cellcolor{green!15}\textbf{66.61} & \cellcolor{green!15}\textbf{84.55} & \cellcolor{green!15}\textbf{53.29} & \cellcolor{green!15}\textbf{47.39} & \cellcolor{green!15}\textbf{82.19} & \cellcolor{green!15}\textbf{63.85} \\
        \bottomrule
    \end{tabular}}
    \caption{Performance of different models on MRB. Each column corresponds to a task category, with the number of test sets indicated in parentheses. Evaluation metrics are provided in Appendix~\ref{ap:main_result}. We adopt GME-2B as the retrieval backbone, while all other models rerank the top-100 retrieved candidates. 
    \textcolor{green!50!black}{\rule{6pt}{6pt}} indicates the best result in reranking models, and 
    \textcolor{gray!70}{\rule{6pt}{6pt}} indicates the second-best.}
    \label{tab:main_result}
    \vspace{-1em}
\end{table*}

\subsection{Settings}\label{sec:exp:setting}
\paragraph{Training Dataset}

To develop a universal multimodal reranking model, we follow the settings of GME and curate training data from three categories: single-modal data (T$\rightarrow$T, I$\rightarrow$I), cross-modal data (I$\leftrightarrow$T, T$\rightarrow$VD), and fused-modal data (IT$\leftrightarrow$T, IT$\rightarrow$I, IT$\rightarrow$IT). In total, we compile approximately \textbf{1.5} million training instances from diverse sources, including M-BEIR \citep{cong2025uniir}, ViDoRe \citep{faysse2025colpali}, ImageNet-1K \citep{5206848}, E-VQA \citep{10376881}, and MS MARCO \citep{DBLP:journals/corr/NguyenRSGTMD16}.
To ensure fairness and efficiency in the comparative experiments reported in \S \ref{sec:analysis}, we additionally construct a balanced and category-representative subset consisting of about 270K samples drawn from the full training dataset. The models, GMR-3B and GMR-7B, are trained on the complete dataset to achieve optimal performance, whereas the models evaluated in \S \ref{sec:analysis} are trained on the constructed subset. 
Details could be found in Appendix \ref{ap:exp:train_dataset}.

\paragraph{MRB Benchmark}

To facilitate a more rigorous evaluation of model performance, we construct the MRB benchmark, which comprises \textbf{40} test datasets sourced from BEIR \citep{kamalloo:2024}, UMRB \citep{Zhang_2025_CVPR}, ViDoRe \citep{faysse2025colpali,macé2025vidorebenchmarkv2raising}, and MIEB \citep{xiao2025miebmassiveimageembedding}. Collectively, these datasets span diverse modalities, domains, and task types, ensuring that the benchmark provides a comprehensive and representative assessment of model generalization. To more clearly highlight performance differences among models, we exclude test datasets on which GME-2B exhibits exceptionally high performance. A detailed description of the MRB benchmark composition is provided in Appendix \ref{ap:exp:test_dataset}.

\paragraph{Training Configuration}
We adopt the Qwen2.5-VL-Instruction \citep{qwen2.5-VL} model series as the backbone of our multimodal large language model (MLLM), and conduct training at both 3-billion (3B) and 7-billion (7B) parameter scales. For efficient adaptation, we employ Low-Rank Adaptation (LoRA) with a rank of 16 and a learning rate of 1e-4. As evidenced by the comparative results in \S \ref{sec:analysis}, within the domain of multimodal LLM reranking, SFT consistently outperforms CL. Consequently, we adopt SFT as the training strategy for our GMR series models.

During training, we set the maximum input length to 3,200 tokens. Each training sample is paired with 16 negative instances for the GMR-3B and GMR-7B models, and with 4 negative instances for the models mentioned in \S \ref{sec:analysis}. Regarding the selection of negatives, we employ two strategies: \emph{Random Selection} and \emph{Hard Mining}, maintaining a balanced ratio of 1:1 between them. Further details on the negative sampling strategy are provided in Appendix \ref{app:exp:neg}. To optimize GPU memory usage, we train the model using bfloat16 precision. All experiments were conducted on eight NVIDIA A100 GPUs, each equipped with 80 GB of memory.

\paragraph{Baselines}
\label{}
We adopt GME-2B as the retrieval backbone to generate candidate results for each task. Specifically, the top-100 retrieved candidates are retained, and all reranking models are subsequently evaluated on this candidate pool. For the experiment described in \S\ref{sec:analysis}, we reorder the top-25 candidates to balance fairness with efficiency.
Our method is compared against three representative types of reranking systems:
(1) A representative textual model : Qwen3-Reranker \citep{zhang2025qwen3e} ({\small{\textit{Qwen3}}}), exemplifying recent advancements in text-based reranking.
(2) A versatile multimodal reranking model: Jina-rerank-m0\footnote{https://huggingface.co/jinaai/jina-reranker-m0}({\small{\textit{Jina-m0}}}).  This model natively supports single-modal tasks and cross-modal tasks. Leveraging the flexibility of its MLLM architecture, we extend its application to fused-modal tasks by adopting its input template. The specifics of these adaptations are detailed in Appendix \ref{app:model_setting:Jina-rerank-m0}.
(3) A cutting-edge visual document reranking model: MonoQwen2-VL-v0.1 \citep{chaffin2024monoqwen} ({\small{\textit{MonoQwen}}}). Similar to our approach with Jina-rerank-m0, we evaluate this model across all task types.  The input templates used is provided in Appendix \ref{app:model_setting:MonoQwen2-VL-v0.1}.

This comprehensive evaluation benchmarks our method against leading models across diverse modalities and task types, enabling a thorough assessment of its effectiveness.

\subsection{Main Results}
\label{sect:main_result}
\begin{wrapfigure}{r}{0.325\textwidth}  % r表示右侧，0.5\textwidth控制宽度
\vspace{-2em}  % 调整垂直间距
\includegraphics[width=0.325\textwidth]{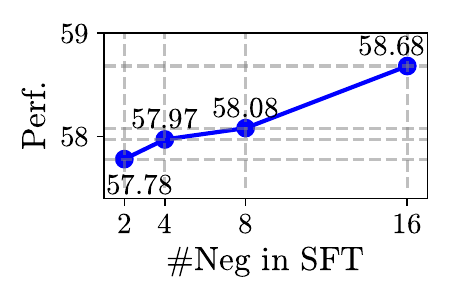}
\vspace{-2em}  % 调整垂直间距
\caption{Results with different numbers of negative in SFT.}
\label{fig:neg_num}
\vspace{-1em}  % 调整垂直间距
\end{wrapfigure}

We first examine the effect of the number of negatives. In SFT, where query–candidate similarity is formulated as a binary classification task, the number of negatives directly affects model performance. To identify an appropriate setting under our computational budget, we experiment with varying numbers of negatives (\Figref{fig:neg_num}). Performance consistently improves with more negatives, peaking at 16. Moreover, SFT outperforms CL across all settings (Appendix \ref{ap:main_result:neg_num}). Based on these results, we set the number of negatives to 16 in training. Given the impact of random initialization on performance (\S \ref{sec:analysis:direction}), we also conduct an ablation on freezing the LM head (Appendix \ref{ap:main_result_nofreeze}) and find that has no effect on SFT performance.

We next examine the evaluation results. Table \ref{tab:main_result} presents a comprehensive overview of the baseline systems' performance. The reported scores are averaged across the respective sub-tasks and are organized according to the retrieval modality: Single-Modal, Cross-Modal, and Fused-Modal. For completeness, the overall micro-average score across all sub-tasks is provided in the final column.

\textbf{Achieve state-of-the-art performance in universal multimodal reranking.} Analyzing the average metrics, our smaller model, GMR-3B, exhibits superior results compared to the fused-modal reranking model (Jina-Rerank-m0). The larger GMR-7B further elevates this performance, underscoring the efficacy in addressing universal multimodal reranking challenges.

\textbf{Rival and surpass leading textual reranker.} We conduct a comparative analysis with the state-of-the-art textual reranking model, Qwen3-Reranker, which is specifically optimized for the T$\rightarrow$T task within the Single-Modal category and comprises approximately 4 billion parameters. Our smaller model exhibited similar performance metrics when evaluated against models of similar parameter scale. Notably, our larger model surpass the performance of Qwen3-Reranker, providing strong empirical evidence for the efficacy of our proposed methodology. 

\textbf{Adapt seamlessly to visual-document reranking.} We compare with the visual document reranking model, MonoQwen2-VL-v0.1, which is specifically tailored for the T$\rightarrow$VD task. Our proposed models demonstrate performance metrics that are surpass those of this task-specific baseline, which suggests a promising direction for developing more efficient and adaptable information re-reanking systems that can seamlessly handle diverse modalities within a single architecture.

\section{Conclusion}

In summary, our study shows that supervised fine-tuning (SFT) consistently outperforms contrastive learning (CL) for LLM-based reranking. By decomposing the loss into \emph{weight} and \emph{direction} components, we find that the weight term primarily drives performance gains by strengthening optimization signals and providing input-specific guidance. While SFT’s directional component is nearly optimal, CL requires learning a score-projection matrix to achieve comparable results. Building on these insights, we develop the GMR-3B and GMR-7B models, which set new state-of-the-art results on the MRB benchmark covering 40 datasets. By releasing MRB, our models, and code, we provide a solid foundation for future research in large-scale multimodal retrieval and universal LLM reranking, underscoring both methodological and practical significance.
\newpage

\bibliography{ref}
\bibliographystyle{iclr2026_conference}

\clearpage

\appendix
\section*{Appendix}

\section{METHOD DETAILS}
\subsection{GMR input template}\label{app:method:gmr_input}
\begin{figure}[H]
\centering
\includegraphics[width=0.8\linewidth]{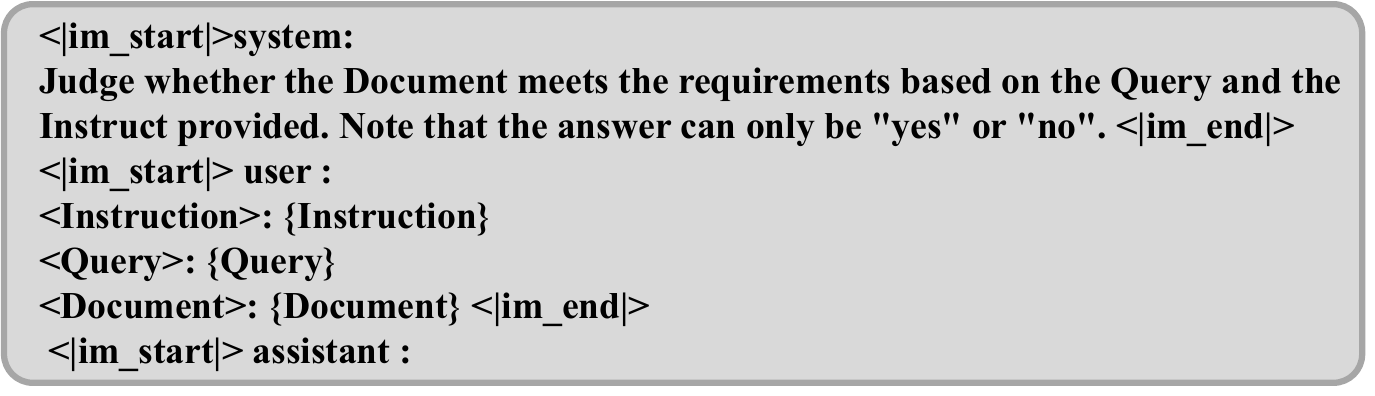}
\caption{The structured input template for GMR series models.}
\label{fig:input_prompt}
\end{figure}

Following a chat-based template, The prompt formulates a binary classification task by providing the model with a specific Instruction, Query, and Document for evaluation as shon in \Figref{fig:input_prompt}.

\subsection{Loss Function Decomposition}
\label{ap:loss_function_decomposition}
In this section, we elaborate on the derivation process of the equation in \S \ref{sec:method:loss_decompose}.
\label{app:LFD}
\begin{itemize}
\item \Eqref{eq:cl_pos}:
\begin{align}
-\frac{\partial\mathcal{L}^\mathrm{CL}}{\partial \mathbf{h}_0^+}
&=-\frac{\partial\mathcal{L}^\mathrm{CL}}{\partial s^y(\mathbf{h}_0^+)}\frac{\partial s^y(\mathbf{h}_0^+)}{\partial \mathbf{h}_0^+}
\nonumber\\
&= -\frac{\partial(-\log\frac{\exp(\sigma(ins, q, d_0^+))}{\exp(\sigma(ins, q, d_0^+))+\sum_i\exp(\sigma(ins, q, d_i^-))})}{\partial s^y(\mathbf{h}_0^+)}\frac{\partial s^y(\mathbf{h}_0^+)}{\partial \mathbf{h}_0^+}
\nonumber\\
&= \frac{\partial(\log\frac{\exp(s^y(\mathbf{h}_0^+))}{\exp(s^y(\mathbf{h}_0^+))+\sum_iexp(s^y(\mathbf{h}^-_i))})}{\partial s^y(\mathbf{h}_0^+)}\frac{\partial s^y(\mathbf{h}_0^+)}{\partial \mathbf{h}_0^+}
\nonumber\\
&=\frac{\exp(s^y(\mathbf{h}_0^+))+\sum_iexp(s^y(\mathbf{h}^-_i))}{\exp(s^y(\mathbf{h}_0^+))}\cdot\frac{\partial(\frac{\exp(s^y(\mathbf{h}_0^+))}{\exp(s^y(\mathbf{h}_0^+))+\sum_iexp(s^y(\mathbf{h}^-_i))})}{\partial s^y(\mathbf{h}_0^+)}\cdot\frac{\partial s^y(\mathbf{h}_0^+)}{\partial \mathbf{h}_0^+}\nonumber\\
&=\frac{\exp(s^y(\mathbf{h}_0^+))+\sum_iexp(s^y(\mathbf{h}^-_i))}{\exp(s^y(\mathbf{h}_0^+))}\cdot\frac{\sum_iexp(s^y(\mathbf{h}^-_i))}{(\exp(s^y(\mathbf{h}_0^+))+\sum_iexp(s^y(\mathbf{h}^-_i)))^2}\nonumber\\
&\quad\cdot\frac{\partial exp(s^y(\mathbf{h}_0^+))}{\partial s^y(\mathbf{h}_0^+)}\cdot\frac{\partial s^y(\mathbf{h}_0^+)}{\partial \mathbf{h}_0^+}\nonumber\\
&=\frac{\exp(s^y(\mathbf{h}_0^+))+\sum_iexp(s^y(\mathbf{h}^-_i))}{\exp(s^y(\mathbf{h}_0^+))}\cdot\frac{\sum_iexp(s^y(\mathbf{h}^-_i))}{(\exp(s^y(\mathbf{h}_0^+))+\sum_iexp(s^y(\mathbf{h}^-_i)))^2}\nonumber\\
&\quad\cdot\exp(s^y(\mathbf{h}_0^+))\cdot\frac{\partial s^y(\mathbf{h}_0^+)}{\partial \mathbf{h}_0^+}\nonumber\\
&=\frac{\sum_j\exp(s^y(\mathbf{h}^-_j))}{\exp(s^y(\mathbf{h}_0^+))+\sum_j\exp(s^y(\mathbf{h}^-_j))}\frac{\partial s^y(\mathbf{h}_0^+)}{\partial \mathbf{h}_0^+} \nonumber\\
&=\frac{\sum_j\exp(s^y(\mathbf{h}^-_j))}{\exp(s^y(\mathbf{h}_0^+))+\sum_j\exp(s^y(\mathbf{h}^-_j))}\mathcal{M}_y
\end{align}

\newpage
\item \Eqref{eq:cl_neg}:
\begin{align}
-\frac{\partial\mathcal{L}^\mathrm{CL}}{\partial \mathbf{h}^-_i}
&=-\frac{\partial\mathcal{L}^\mathrm{CL}}{\partial s^y(\mathbf{h}^-_i)}\frac{\partial s^y(\mathbf{h}^-_i)}{\partial \mathbf{h}^-_i}\nonumber\\
&=-\frac{\partial(-\log\frac{\exp(\sigma(ins, q, d_0^+))}{\exp(\sigma(ins, q, d_0^+))+\sum_i\exp(\sigma(ins, q, d_i^-))})}{\partial s^y(\mathbf{h}_i^-)}\frac{\partial s^y(\mathbf{h}^-_i)}{\partial \mathbf{h}^-_i}\nonumber\\
&= \frac{\partial(\log\frac{\exp(s^y(\mathbf{h}_0^+))}{\exp(s^y(\mathbf{h}_0^+))+\sum_iexp(s^y(\mathbf{h}^-_i))})}{\partial s^y(\mathbf{h}_i^-)}\frac{\partial s^y(\mathbf{h}_i^-)}{\partial \mathbf{h}_i^-}\nonumber\\
&=\frac{\exp(s^y(\mathbf{h}_0^+))+\sum_iexp(s^y(\mathbf{h}^-_i))}{\exp(s^y(\mathbf{h}_0^+))}\cdot(-\frac{exp(s^y(h^+_0))}{(\exp(s^y(\mathbf{h}_0^+))+\sum_iexp(s^y(\mathbf{h}^-_i)))^2})\nonumber\\
&\quad\cdot\exp(s^y(\mathbf{h}_i^-))\cdot\frac{\partial s^y(\mathbf{h}_i^-)}{\partial \mathbf{h}_0^+}\nonumber\\
&=-\frac{\exp(s^y(\mathbf{h}^-_i))}{\exp(s^y(\mathbf{h}_0^+))+\sum_j\exp(s^y(\mathbf{h}^-_i))}\frac{\partial s^y(\mathbf{h}^-_i)}{\partial \mathbf{h}^-_i} \nonumber\\
&=-\frac{\exp(s^y(\mathbf{h}^-_i))}{\exp(s^y(\mathbf{h}_0^+))+\sum_j\exp(s^y(\mathbf{h}^-_i))}\mathcal{M}_y
\end{align}

\item \Eqref{eq:sft_pos}:
    \begin{align}
-\frac{\partial\mathcal{L}^\mathrm{SFT}}{\partial \mathbf{h}_0^+}
=&-\frac{\partial\mathcal{L}^\mathrm{SFT}}{\partial s^y(\mathbf{h}_0^+)}\frac{\partial s^y(\mathbf{h}_0^+)}{\partial \mathbf{h}_0^+}-\frac{\partial\mathcal{L}^\mathrm{SFT}}{\partial s^n(\mathbf{h}_0^+)}\frac{\partial s^n(\mathbf{h}_0^+)}{\partial \mathbf{h}_0^+}\nonumber\\
=&-\frac{\partial(-\log(p(\text{``yes''}|P(\{\text{``yes'',``no''}\}|\{ins,q,d_i\}))))}{\partial s^y(\mathbf{h}_0^+)}\frac{\partial s^y(\mathbf{h}_0^+)}{\partial \mathbf{h}_0^+}\nonumber\\
&-\frac{\partial(-\log(p(\text{``yes''}|P(\{\text{``yes'',``no''}\}|\{ins,q,d_i\}))))}{\partial s^n(\mathbf{h}_0^+)}\frac{\partial s^n(\mathbf{h}_0^+)}{\partial \mathbf{h}_0^+}\nonumber\\
=&\frac{\partial(\log\frac{e^{P(\text{``yes''} |\{ins, q, d\})}}{e^{P(\text{``yes''} |\{ins, q, d\})}+e^{P(\text{``no''} |\{ins, q, d\})}})}{\partial s^y(\mathbf{h}_0^+)}\frac{\partial s^y(\mathbf{h}_0^+)}{\partial \mathbf{h}_0^+}+\nonumber \\
&\frac{\partial(\log\frac{e^{P(\text{``yes''} |\{ins, q, d\})}}{e^{P(\text{``yes''} |\{ins, q, d\})}+e^{P(\text{``no''} |\{ins, q, d\})}})}{\partial s^n(\mathbf{h}_0^+)}\frac{\partial s^n(\mathbf{h}_0^+)}{\partial \mathbf{h}_0^+}\nonumber\\
=& \frac{\partial(\log\frac{exp(s^y(\mathbf{h}_0^+))}{exp(s^y(\mathbf{h}_0^+))+exp(s^n(\mathbf{h}_0^+))})}{\partial s^y(\mathbf{h}_0^+)}\frac{\partial s^y(\mathbf{h}_0^+)}{\partial \mathbf{h}_0^+}\nonumber\\
&+\frac{\partial(\log\frac{exp(s^y(\mathbf{h}_0^+))}{exp(s^y(\mathbf{h}_0^+))+exp(s^n(\mathbf{h}_0^+))})}{\partial s^n(\mathbf{h}_0^+)}\frac{\partial s^n(\mathbf{h}_0^+)}{\partial \mathbf{h}_0^+}\nonumber\\
=&\frac{exp(s^n(\mathbf{h}_0^+))}{exp(s^y(\mathbf{h}_0^+))+exp(s^n(\mathbf{h}_0^+))}\frac{\partial s^y(\mathbf{h}_0^+)}{\partial \mathbf{h}_0^+} \nonumber\\
&-\frac{exp(s^n(\mathbf{h}_0^+))}{exp(s^y(\mathbf{h}_0^+))+exp(s^n(\mathbf{h}_0^+))}\frac{\partial s^n(\mathbf{h}_0^+)}{\partial \mathbf{h}_0^+} \nonumber\\
=&\frac{exp(s^n(\mathbf{h}_0^+))}{exp(s^y(\mathbf{h}_0^+))+exp(s^n(\mathbf{h}_0^+))}(\frac{\partial s^y(\mathbf{h}_0^+)}{\partial \mathbf{h}_0^+}-\frac{\partial s^n(\mathbf{h}_0^+)}{\partial \mathbf{h}_0^+}) \nonumber\\
=&\frac{\exp(s^n(\mathbf{h}_0^+))}{\exp(s^y(\mathbf{h}_0^+))+\exp(s^n(\mathbf{h}_0^+))}(\mathcal{M}_\mathrm{y}-\mathcal{M}_\mathrm{n})
\end{align}
\newpage
\item \Eqref{eq:sft_neg}:
    
\begin{align}
-\frac{\partial\mathcal{L}^\mathrm{SFT}}{\partial \mathbf{h}^-_i}
=&-\frac{\partial\mathcal{L}^\mathrm{SFT}}{\partial s^y(\mathbf{h}^-_j)}\frac{\partial s^y(\mathbf{h}^-_i)}{\partial \mathbf{h}^-_i}-\frac{\partial\mathcal{L}^\mathrm{SFT}}{\partial s^n(\mathbf{h}^-_i)}\frac{\partial s^n(\mathbf{h}^-_i)}{\partial \mathbf{h}^-_i}\nonumber\\
=&-\frac{\partial(-\log(p(\text{``no''}|P(\{\text{``yes'',``no''}\}|\{ins,q,d_i\}))))}{\partial  s^y(\mathbf{h}^-_i)}\frac{\partial  s^y(\mathbf{h}^-_i)}{\partial \mathbf{h}^-_i}\nonumber\\
&-\frac{\partial(-\log(p(\text{``no''}|P(\{\text{``yes'',``no''}\}|\{ins,q,d_i\}))))}{\partial s^n(\mathbf{h}^-_i)}\frac{\partial s^n(\mathbf{h}^-_i)}{\partial \mathbf{h}^-_i}\nonumber\\
=&\frac{\partial(\log\frac{e^{P(\text{``no''} |\{ins, q, d\})}}{e^{P(\text{``yes''} |\{ins, q, d\})}+e^{P(\text{``no''} |\{ins, q, d\})}})}{\partial s^y(\mathbf{h}^-_i)}\frac{\partial s^y(\mathbf{h}^-_i)}{\partial \mathbf{h}^-_i}+\nonumber \\
&\frac{\partial(\log\frac{e^{P(\text{``no''} |\{ins, q, d\})}}{e^{P(\text{``yes''} |\{ins, q, d\})}+e^{P(\text{``no''} |\{ins, q, d\})}})}{\partial s^n(\mathbf{h}^-_i)}\frac{\partial s^n(\mathbf{h}^-_i)}{\partial \mathbf{h}^-_i}\nonumber\\
=& \frac{\partial(\log\frac{exp(s^n(\mathbf{h}^-_i))}{exp(s^y(\mathbf{h}^-_i))+exp(s^n(\mathbf{h}^-_i))})}{\partial s^y(\mathbf{h}^-_i)}\frac{\partial s^y(\mathbf{h}^-_i)}{\partial \mathbf{h}^-_i}\nonumber\\
&+\frac{\partial(\log\frac{exp(s^n(\mathbf{h}^-_i))}{exp(s^y(\mathbf{h}^-_i))+exp(s^n(\mathbf{h}^-_i))})}{\partial s^n(\mathbf{h}^-_i)}\frac{\partial s^n(\mathbf{h}^-_i)}{\partial \mathbf{h}^-_i}\nonumber\\
=&-\frac{\exp(s^y(\mathbf{h}^-_i))}{\exp(s^y(\mathbf{h}^-_i))+\exp(s^n(\mathbf{h}^-_i))}\nonumber(\frac{\partial s^y(\mathbf{h}^-_i)}{\partial \mathbf{h}^-_i}-\frac{\partial s^n(\mathbf{h}^-_i)}{\partial \mathbf{h}^-_i})\nonumber\\
=&-\frac{\exp(s^y(\mathbf{h}^-_i))}{\exp(s^y(\mathbf{h}^-_i))+\exp(s^n(\mathbf{h}^-_i))}(\mathcal{M}_y-\mathcal{M}_n)
\end{align}
    
\end{itemize}

% \newpage
\section{Analysis Experiment}
\subsection{The influence of precision on CL}
\begin{wraptable}{r}{0.4\textwidth}
\setlength{\tabcolsep}{3pt}
\vspace{-1em}  % 调整垂直间距
\begin{tabular}{cccc}
    \toprule
    \textbf{Method}      & \textbf{Precision} & \textbf{Avg}  & \textbf{$\Delta$} \\
    \midrule
    \multirow{2}{*}{CL} & FP16   & 56.09 & -    \\
     & FP32 & 56.40 & {\textcolor{green}{$\blacktriangle$}~0.31} \\
    \bottomrule
\end{tabular}
\caption{Impact of precision on Contrastive Learning's performance.}
\vspace{-1em}
\label{ap:analysis:cl_precision}
\end{wraptable}
We validate the findings of FlatNCE by performing full half-precision training during loss function computation on the contrastive learning (CL) model. Specifically, we configure the model to use BF16 for accuracy, and in the loss computation process (refer to \Algref{alg:wd-loss}), we control all other variables while varying the precision of the weight computations between FP16 and FP32 to assess their impact on model performance. The results show that FP32 precision yields better performance than FP16 precision, confirming that computational precision significantly affects the effectiveness of contrastive learning.

\subsection{Function of weight}
\label{app:analysis:weight}
\begin{wraptable}{r}{0.335\textwidth}
\vspace{-1.5em}  % 调整垂直间距
\begin{tabular}{ccc}
    \toprule
    \textbf{Method}      & \textbf{$\tau$} & \textbf{Avg} \\
    \midrule
    \multirow{3}{*}{w/ $\tau$ mask} & 1e-2   & 55.07   \\
     & 1e-3 & 56.57 \\
     & 1e-4 & 55.89 \\
    \bottomrule
\end{tabular}
\caption{The performance of the model under different values of $\tau$.}
\vspace{-2em}
\label{ap:analysis:w_tau}
\end{wraptable}
To investigate the role of the weight, we first define $s (h_i) = \frac{exp (s ^ y (h_i))} {exp (s ^ y (h_i)) + exp (s ^ n (h_i))}$. 
Since $s(h_i)$ is bounded within $[0,1]$, prior experience with embedding models suggests that an appropriate scaling factor is necessary to accelerate model convergence. Therefore, we introduce a temperature parameter $\beta = 5 \times 10^{-2}$ into \Eqref{function:w_base}, yielding $W^-_j=\frac{exp(s(\mathbf{h}^-_j)/\beta)}{\sum_j exp(s(\mathbf{h}^-)/\beta)}$. 
In addition, for experiments involving the masking rule, we vary $\tau \in {10^{-2}, 10^{-3}, 10^{-4}}$ to identify the configuration that achieves optimal performance.
For the experiment with $W_{CL}$, we follow \Eqref{eq:cl_pos_weight} and \ref{eq:cl_neg_weight}, consistent with the requirements of contrastive learning, where the positive and negative weights must satisfy the constraint $W^+ = \sum W^-$. Since directly setting $W_{+W_{CL}} = W_{Base}W_{CL}$ would violate this condition, we instead use $W_{+W_{CL}} = W_{CL}$ for comparison with $W_{Base}$.
For the experiment with $W_{SFT}$, we aim to demonstrate that $W_{SFT}$ can effectively enhance the performance of $W_{Base}$. Following \Eqref{eq:sft_pos_weight} and \ref{eq:sft_neg_weight}, we set $W_{+W_{SFT}} = W_{Base} W_{SFT}$ and evaluate its impact on model performance.

\subsection{The influence of Token selection}
\label{app:analysis:token}
To examine whether introducing additional tokens during training can enhance the directionality component and improve model performance, we randomly sample 10,000 instances together with their corresponding positives and negatives. Based on the model outputs, we identify the top 16 tokens with the highest average logits, which include “yes” and “no.” The remaining tokens in this set are: $\{$“No,” “Yes,” “NO,” “YES,” “The,” “None,” “In,” “Answer,” “This,” “To,” “Not,” “not,” “There,” “-no”$\}$.

\section{Experiment Setting}
\label{ap:exp}
\subsection{Training Datasets}
\label{ap:exp:train_dataset}

\begin{wrapfigure}{r}{0.45\textwidth}  % r表示右侧，0.5\textwidth控制宽度
\vspace{-2em}  % 调整垂直间距
\includegraphics[width=0.45\textwidth]{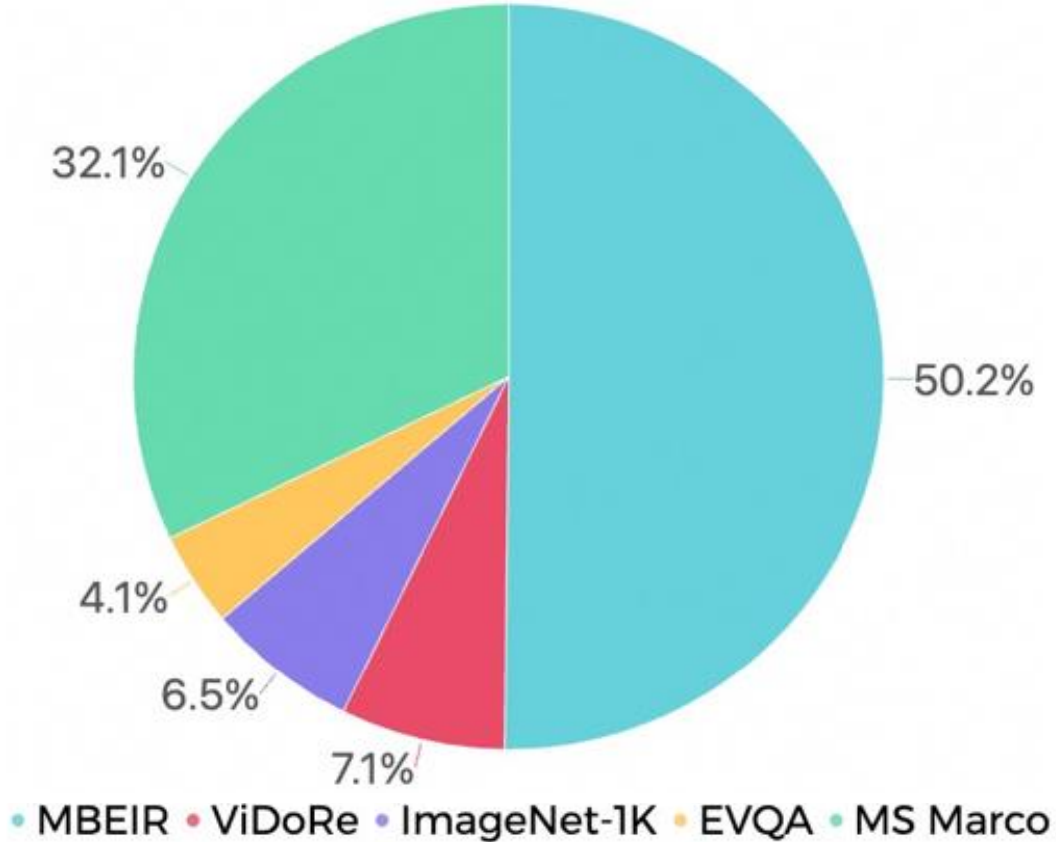}
\vspace{-1em}  % 调整垂直间距
\caption{The proportion of the training data.}
\label{fig:train_data}
\vspace{-2em}  % 调整垂直间距
\end{wrapfigure}
Our training dataset is curated from diverse sources, including M-BEIR, ViDoRe, ImageNet-1K, E-VQA, and Ms Marco. These datasets cover a wide array of domains, ensuring that the model is exposed to varied and representative examples across different tasks. To ensure balanced representation across task domains, we sample 100k instances from ImageNet-1K and integrated them into our training corpus.

In total, our training dataset consists of approximately 1.5 million instances, which are distributed across various domains to ensure robust learning. The detailed distribution of the data across these domains is carefully visualized in \Figref{fig:train_data}.

To ensure a fair comparison between supervised fine-tuning and contrastive learning, we
construct a balanced, category-representative subset of approximately 270K samples from our training dataset, and the details could be found in Table \ref{tab:sub_trainset}.

\begin{table}[htbp]
\centering
\resizebox{0.95\linewidth}{!}{
\begin{tabular}{l l l r}
\toprule
\textbf{Class} & \textbf{Task} & \textbf{Datasets} & \textbf{Number} \\
\midrule
\multirow{2}{*}{Single-Modal(4)} & \multirow{1}{*}{T$\rightarrow$T (2)} & WebQA$\dagger$ Ms Marco & \multirow{1}{*}{30000} \\
\cmidrule(l){2-4}
& \multirow{1}{*}{I$\rightarrow$I (2)} & Nights$\dagger$ ImageNet-1K & \multirow{1}{*}{30000} \\
\midrule
\multirow{3}{*}{Cross-Modal(6)} & \multirow{1}{*}{T$\rightarrow$I (2)} & Fashion200k$\dagger$ VisualNews$\dagger$ & \multirow{1}{*}{29958} \\
\cmidrule(l){2-4}
& T$\rightarrow$VD (1) & ViDoRe  & 30000 \\
\cmidrule(l){2-4}
& \multirow{1}{*}{I$\rightarrow$T (3)} & Fashion200k$\dagger$ MScoco$\dagger$ VisualNews$\dagger$ & \multirow{1}{*}{30882} \\
\midrule
\multirow{4}{*}{Fused-Modal(11)} & \multirow{1}{*}{T$\rightarrow$IT (2)} & EDIS$\dagger$ WebQA$\dagger$ & \multirow{1}{*}{30000} \\
\cmidrule(l){2-4}
& \multirow{1}{*}{IT$\rightarrow$T (3)} & LLava$\dagger$ OVEN$\dagger$ Remuq$\dagger$ & \multirow{1}{*}{30382} \\
\cmidrule(l){2-4}
& \multirow{1}{*}{IT$\rightarrow$I (2)} & CIRR$\dagger$ FashionIQ$\dagger$ & \multirow{1}{*}{29528} \\
\cmidrule(l){2-4}
& \multirow{1}{*}{IT$\rightarrow$IT (3)} & E-VQA OVEN$\dagger$ & \multirow{1}{*}{30000} \\
\bottomrule
\end{tabular}
}
\caption{The details of sub trainset. $\dagger$ means that they belong to the M-BEIR dataset. }
\label{tab:sub_trainset}
\vspace{-1em}
\end{table}

\subsection{MRB Benchmark}
\label{ap:exp:test_dataset}
Since overly simple tasks fail to effectively differentiate the performance of various rerank models, we exclude the dataset on which the GME-2B model achieves exceptionally high performance. Detailed descriptions of MRB Benchmark are provided in Tables \ref{tab:test_datasets} and \ref{tab:benchmark_detail}.
\begin{table}[htbp]
\centering
\resizebox{0.9\linewidth}{!}{
\begin{tabular}{lll}
\toprule
\textbf{Class} & \textbf{Task} & \textbf{Datasets} \\
\midrule
\multirow{4}{*}{Single-Modal(15)} & \multirow{3}{*}{T$\to$T (14)} & ArguAna$\dagger$ Climate-FEVER$\dagger$ CQADupStack$\dagger$ DBPedia$\dagger$ \\
& & FIQA2018$\dagger$ HotpotQA$\dagger$  MSMARCO$\dagger$ NFCorpus$\dagger$ NQ$\dagger$\\
& & Quora$\dagger$ SCIDOCS$\dagger$ SciFact$\dagger$ Touche2020$\dagger$ TRECCOVID$\dagger$\\ \cmidrule(l){2-3}
& I$\to$I (1) & Nights$^{*}$ \\
\midrule
\multirow{5}{*}{Cross-Modal(14)} & \multirow{1}{*}{T$\to$I (4)} & VisualNews$^{*}$ Fashion200k$^{*}$ Memotion$^{\star}$ HatefulMemes$^{\star}$\\ \cmidrule(l){2-3}
& \multirow{2}{*}{T$\to$VD (5)} & TAT-DQA$^{\dag}$ ArxivQA$^{\dag}$ DocVQA$^{\dag}$ \\
& & MIT Tissue Interaction$^{\dag}$ World Economic Reports$^{\dag}$ \\ \cmidrule(l){2-3}
& \multirow{2}{*}{I$\to$T (5)} & VisualNews$^{*}$ Fashion200K$^{*}$ \\
& & Memotion$^{\star}$ GLDv2$^{\star}$ HatefulMemes$^{\star}$\\
\midrule
\multirow{5}{*}{Fused-Modal(11)} & T$\to$IT (2) & WebQA$^{*}$ EDIS$^{*}$ \\ \cmidrule(l){2-3}
& \multirow{1}{*}{IT$\to$T (4)} & OVEN$^{*}$ INFOSEEK$^{*}$  OKVQA$^{*}$ VizWiz$^{\star}$\\ \cmidrule(l){2-3}
& IT$\to$I (2) & FashionIQ$^{*}$ CIRR$^{*}$ \\ \cmidrule(l){2-3}
& IT$\to$IT (3) & OVEN$^{*}$ E-VQA$^{*}$ INFOSEEK$^{*}$ \\
\bottomrule
\end{tabular}}
\caption{An overview of datasets in \emph{MRB}. $\dagger$ means it belong to BEIR. $^{*}$ means it belong to UMRB. $^{\dag}$ means it belong to ViDoRe. $^{\star}$ means it belong to MIEB.}
\label{tab:test_datasets}
\vspace{-1.5em}
\end{table}

\begin{table*}[!t]
\centering
\scriptsize
\resizebox{0.9\textwidth}{!}{\begin{tabular}{l|ccccccc}
\toprule
\bf Name & \bf Type & \bf Categ. & \bf Eval & \bf Candidates & \bf Eval Query & \bf Eval Candidate \\
& & & \bf Samples & \bf Nums & \bf avg. chars & \bf avg. chars \\
\midrule
\midrule
ArguAna & Single-Modal & T$\rightarrow$T & 1406 & 8,674 & 192.98 & 166.80  \\
Climate-FEVER & Single-Modal & T$\rightarrow$T & 1,535 & 5,416,593 & 20.13 &  84.76  \\
CQADupStack & Single-Modal & T$\rightarrow$T & 13,145 & 457,199 & 8.59 & 129.09  \\
DBPedia & Single-Modal & T$\rightarrow$T & 400 & 4,635,922 & 5.39 & 49.68  \\
FiQA2018 & Single-Modal & T$\rightarrow$T & 648 & 57,638 & 10.77 & 132.32  \\
HotpotQA & Single-Modal & T$\rightarrow$T & 7,405 & 5,233,329 & 17.61 & 46.30  \\
MSMARCO & Single-Modal & T$\rightarrow$T & 6,980 & 8,841,823 & 5.96 &  55.98  \\
NFCorpus & Single-Modal & T$\rightarrow$T & 323 & 3,633 & 3.30 & 232.26  \\
NQ & Single-Modal & T$\rightarrow$T & 3,452 & 2,681,468 & 9.16 & 78.88  \\
Quora & Single-Modal & T$\rightarrow$T & 10,000 & 522,931 & 9.53 & 11.44  \\
SCIDOCS & Single-Modal & T$\rightarrow$T & 1,000 & 25,657 & 9.38 & 176.19  \\
SciFact & Single-Modal & T$\rightarrow$T & 300 & 5,183 & 12.37 & 213.63  \\
Touche2020 & Single-Modal & T$\rightarrow$T & 49 & 382,545 & 6.55 & 292.37   \\
TRECCOVID & Single-Modal & T$\rightarrow$T & 50 & 171,332 & 10.60 & 160.77  \\
Nights & Single-Modal & I$\rightarrow$I  & 2,120 & 40,038 & - & -  \\
\midrule
VisualNews & Cross-Modal & T$\rightarrow$I  & 19,995 & 542,246 & 18.78 & -   \\
Fashion200k & Cross-Modal & T$\rightarrow$I  & 1,719 & 201,824 & 4.89 & -  \\
HatefulMemes & Cross-Modal & T$\rightarrow$I  & 1000 & 10000 & 10.42 & - 
\\
Memotion & Cross-Modal & T$\rightarrow$I  & 697 & 6988 & 14.77 & - 
\\
TAT-DQA & Cross-Modal & T$\rightarrow$VD  & 1,646 & 277 & 12.44 & -   \\
ArxivQA & Cross-Modal & T$\rightarrow$VD  & 500 & 500 & 17.12 & -  \\
DocVQA & Cross-Modal & T$\rightarrow$VD  & 451 & 500 & 8.23 & -   \\
WER & Cross-Modal & T$\rightarrow$VD  & 58 & 452 & 13.05 & -  \\
MITTI & Cross-Modal & T$\rightarrow$VD  & 160 & 1016 & 13.91 & -  \\

VisualNews & Cross-Modal & I$\rightarrow$T  & 20,000 & 537,568 & - & 18.53  \\
Fashion200k & Cross-Modal & I$\rightarrow$T  & 4,889 & 61,707 & - & 4.95   \\
GLDv2 & Cross-Modal & I$\rightarrow$T  & 1704 & 674 & - & 3.18 \\
Memotion & Cross-Modal & T$\rightarrow$I  & 697 & 6988 & - & 14.67 \\
HatefulMemes & Cross-Modal & I$\rightarrow$T  & 1000 & 10000 & - & 11.53 \\
\midrule

WebQA & Fused-Modal & T$\rightarrow$IT & 2,511 & 403,196 & 16.43 & 12.83  \\
EDIS & Fused-Modal & T$\rightarrow$IT & 3,241 & 1,047,067 & 20.07 & 15.53  \\
OVEN & Fused-Modal & IT$\rightarrow$T &  50,004 & 676,667 & 6.52 & 82.13   \\
INFOSEEK & Fused-Modal & IT$\rightarrow$T & 11,323 & 611,651 & 8.76 & 91.49  \\
OKVQA & Fused-Modal & IT$\rightarrow$T & 5,046 & 114,516 & 8.09 & 102.55   \\
VizWiz & Fused-Modal & IT$\rightarrow$T & 4319 & 2091 & 7.17 & -   \\

FashionIQ & Fused-Modal & IT$\rightarrow$I & 6,003 & 74,381 & 11.70 & -  \\
CIRR & Fused-Modal & IT$\rightarrow$I & 4,170 & 21,551 & 11.01 & -   \\

OVEN & Fused-Modal & IT$\rightarrow$IT & 14,741 & 335,135 & 5.91 & 94.76  \\
EVQA & Fused-Modal & IT$\rightarrow$IT & 3,743 & 68,313 & 9.38 & 211.12  \\
INFOSEEK & Fused-Modal & IT$\rightarrow$IT & 17,593 & 481,782 & 7.94 & 96.00  \\
\bottomrule
\end{tabular}}
\caption{Tasks in \emph{MRB}. Following UMRB, We count the number of datasets under each task type, the number of evaluation instances, the size of the candidate set, and the average length of the text.}
\label{tab:benchmark_detail}
\vspace{-1em}
\end{table*}

\subsection{Negative Selection}\label{app:exp:neg}
The quality and diversity of negatives greatly affect the final performance of the reranker.
Overly simple negatives can make the model lack the ability to distinguish hard negatives from positives, while overly difficult documents are very likely to be false negatives that give the model incorrect update signal.
Therefore, we adopt two strategies to select negatives:
\textbf{(1) Random Selection}. Randomly select irrelevant document as negatives to enhance the generalization ability of the model.
\textbf{(2) Hard Mining}. For each query in every dataset, we use GME-2B to search for the corresponding documents to obtain the top 100, and randomly select $k$ irrelevant samples from them as hard negatives to improve the reranking performance.
We employ this set of hard negatives for all the models trained in this paper.
While training, we always maintain the ratio of random negatives to hard negatives at 1:1 to balance the diversity and quality of the data.

\section{Model Settings}
\label{app:model_setting}
\subsection{GME-2B}
\label{app:model_setting:GME-2B}
We employ the GME-2B model as the foundational retrieval model, generating the initial retrieval results that serve as the input to our diverse reranking approaches. Recognizing that the GME series models leverage instruction fine-tuning, we incorporate task-specific instructions into the input query to enhance the retrieval model's performance.

Aligning with the UMRB benchmark, we curate the specific instructions for each task, as comprehensively detailed in Table \ref{tab:task_instruction}.

\subsection{Qwen3-Reranker}
\label{app:model_setting:Qwen3-Reranker}
Paralleling our approach, Qwen3-Reranker leverages Large Language Models for point-wise reranking within a singular contextual framework. To facilitate instruction-following capabilities, the model incorporates task-specific instructions directly into the input context. By utilizing the LLM's inherent chat template, the similarity assessment is reframed as a binary classification paradigm.

Specifically, for $T\to T$ tasks, we set task-specific instructions the same as GME, as comprehensively illustrated in Table \ref{tab:task_instruction}.
\subsection{GMR}
\label{app:model_settings:GMR}
In our GMR series models, we incorporate the retrieval instructions into the input context, yielding two advantages. Primarily, this approach eliminates the need for task-specific instruction redesign at the reranking stage, enabling seamless instruction transfer from the retrieval phase.

Moreover, by strategically integrating instructions into the contextual input, we effectively guide the model's comprehension, facilitating enhanced task understanding and robust instruction-following capabilities. The comprehensive instruction sets for both training and testing phases are meticulously detailed in Tables \ref{tab:trainset_instruction} and \ref{tab:task_instruction}, respectively.
\begin{table*}[htbp]
\centering
\scriptsize
\setlength{\tabcolsep}{4pt}
\resizebox{0.95\linewidth}{!}{
\begin{tabular}{@{}lll@{}}
\toprule
\bf Task& \bf Dataset & \multicolumn{1}{c}{\bf Query Instruction} \\
\midrule

\multirow{2}{*}{T$\rightarrow$T}
&WebQA &  Retrieve passages from Wikipedia that provide answers to the following question.  \\ 
&Ms Marco & Given a question, retrieve relevant passages that answer the question. \\ 
\midrule

\multirow{2}{*}{I$\rightarrow$I}
&Nights & Find a day-to-day image that looks similar to the provided image.  \\
&ImageNet-1K & Retrieve images of the same type as the one in the question. \\
\midrule

\multirow{2}{*}{T$\rightarrow$I}
&Fashion200k & Based on the following fashion description, retrieve the best matching image.  \\ 
&VisualNews & Identify the news-related image in line with the described event.  \\ 
\midrule

\multirow{1}{*}{T$\rightarrow$VD}
&ViDoRe & Find a screenshot that relevant to the user's question. \\ 
\midrule

\multirow{3}{*}{I$\rightarrow$T}
&VisualNews & Find a caption for the news in the given photo.  \\ 
&Fashion200k & Find a product description for the fashion item in the image.\\ 
&MSCOCO & Find an image caption describing the following everyday image.\\ 
\midrule 

\multirow{2}{*}{T$\rightarrow$IT}
&WebQA & Find a Wikipedia image that answers this question.  \\ 
&EDIS & Find a news image that matches the provided caption.  \\
\midrule

\multirow{3}{*}{IT$\rightarrow$T}
&OVEN & Retrieve a Wikipedia paragraph that provides an answer to the given query about the image.   \\ 
&LLava &  Provide a specific decription of the image along with the following question.  \\ 
&Remuq & Retrieve a fact-based paragraph that provides an answer to the given query about the image.  \\ 
\midrule

\multirow{2}{*}{IT$\rightarrow$I}
&FashionIQ & Find a fashion image that aligns with the reference image and style note.  \\ 
&CIRR &  Retrieve a day-to-day image that aligns with the modification instructions of the provided image. \\ 
\midrule

\multirow{2}{*}{IT$\rightarrow$IT}
&OVEN & Retrieve a Wikipedia image-description pair that provides evidence for the question of this image. \\ 
&E-VQA & Determine the Wikipedia image-snippet pair that matches my question about this image.   \\

\bottomrule
\end{tabular}}
\caption{The instructions for training dataset. We set the instructions for the GMR series models on each task during training as shown in the Table.}
\label{tab:trainset_instruction}
\vspace{-1em}
\end{table*}

\subsection{Jina-rerank-m0}
\label{app:model_setting:Jina-rerank-m0}
Jina-rerank-m0 demonstrates inherent capabilities for processing single-modal and cross-modal tasks. By leveraging the architectural flexibility of Multimodal Large Language Model framework, we extend its operational scope to encompass fused-modal tasks through a input template adaptation.

For text and image-modal inputs, Jina-rerank-m0 organizes Query/Document configurations, as comprehensively illustrated in Table \ref{tab:jina_input}. Building upon this foundational template, we design a input organization strategy for fused-modal scenarios, represented in the \textbf{Fused} configuration.

Ultimately, the model's input is standardized to the canonical format:``$\{\textbf{Document}\}$\texttt{\textbackslash n}$\{\textbf{Query}\}$''.
% \begin{table}[htbp]
% \centering
% \begin{tabular}{@{}llll@{}}
% \toprule
%  & \textbf{Text} & \textbf{Image} & \textbf{Fused} \\
% \midrule
% \textbf{Query} & \text{**Query**:}$\textbackslash$n\{query\} & **Query**:$\textbackslash$n<|vision_start|><|image_pad|><|vision_end|> & **Query**:$\textbackslash$n <|vision_start|><|image_pad|><|vision_end|>\{query\} \\
% \textbf{Document} & **Document**:$\textbackslash$n\{doc\} & **Document**:$\textbackslash$n<|vision_start|><|image_pad|><|vision_end|> & **Document**:$\textbackslash$n <|vision_start|><|image_pad|><|vision_end|>\{doc\} \\
% \bottomrule
% \end{tabular}
% \caption{The input template of Jina-rerank-m0.}
% \label{tab:jina_input}
% \end{table}
\begin{table*}[htbp]
\centering
% l = left-aligned column. @{} removes extra space at the edges.
\resizebox{0.95\linewidth}{!}{
\begin{tabular}{@{}lll@{}}
\toprule
 & \textbf{Query} & \textbf{Document} \\
\midrule
\multirow{2}{*}{\textbf{Text}} 
  & \texttt{**Query**:\textbackslash n\{query\}} 
  & \texttt{**Document**:\textbackslash n \{doc\}} \\ 
\addlinespace % Adds a little extra vertical space for readability
\multirow{2}{*}{\textbf{Image}}
  & \texttt{**Query**:} 
  & \texttt{**Document**:} \\
  & \texttt{\textless vision\_start\textgreater\textless image\_pad\textgreater\textless vision\_end\textgreater} 
  & \texttt{\textless vision\_start\textgreater\textless image\_pad\textgreater\textless vision\_end\textgreater} \\
\addlinespace
\multirow{2}{*}{\textbf{Fused}}
  & \texttt{**Query**:}
  & \texttt{**Document**:} \\
  & \texttt{\textless vision\_start\textgreater\textless image\_pad\textgreater\textless vision\_end\textgreater\{query\}} 
  & \texttt{\textless vision\_start\textgreater\textless image\_pad\textgreater\textless vision\_end\textgreater\{doc\}} \\
\bottomrule
\end{tabular}}
% \caption{The input template of Jina-rerank-m0. We refer to the model's format settings for \textbf{Text} and \textbf{Image} to set the input format of fused-modal data, then format the input as ``$\{\textbf{Document}\}$\texttt{\textbackslash n}$\{\textbf{Query}\}$''.}
\caption{The input template of Jina-rerank-m0. We refer to it's format settings for \textbf{Text} and \textbf{Image} to set the input format of fused-modal data, then format the input as ``$\{\textbf{Document}\}$\texttt{\textbackslash n}$\{\textbf{Query}\}$''.}
\vspace{-1em}
\label{tab:jina_input}
\end{table*}

\subsection{MonoQwen2-VL-v0.1}
\label{app:model_setting:MonoQwen2-VL-v0.1}
Analogous to our method approach with Jina-rerank-m0, we conduct a comprehensive evaluation of MonoQwen2-VL-v0.1 across the full spectrum of task types. Given that MonoQwen2-VL-v0.1 is exclusively trained and tested on the T→VD task, its input configuration is specifically tailored to this particular scenario, as illustrated in Table \ref{tab:mono_input}.

Notably, since MonoQwen2-VL-v0.1 does not incorporate additional instructions during training and lacks inherent instruction-following capabilities, we leverage the established T→VD input template to uniformly configure the inputs for all other tasks, as shown under the \textbf{Others} in Table \ref{tab:mono_input}.
\begin{table*}[htbp]
\centering
\vspace{-1em}
% 'l' 列左对齐，'p{width}' 列为段落列，可自动换行。
% 这里的宽度设置为页面文本宽度的75%，您可以根据需要调整。
\scriptsize
\begin{tabular}{lp{0.8\textwidth}}
\toprule
 & \textbf{Input Format} \\
\midrule
$\mathbf{T \to VD}$ & \texttt{\{doc\}\textbackslash nAssert the relevance of the previous image document to the following query, answer True or False. The query is: \{query\}} \\
\addlinespace % 增加行间距以提高可读性
\textbf{Others} & \texttt{\{doc\}\textbackslash nAssert the relevance of the previous document to the following query, answer True or False. The query is: \{query\}} \\
\bottomrule
\end{tabular}
\caption{The input template of MonoQwen2-VL-v0.1. $\mathbf{T \to VD}$ is the original input format of it, and we design the input formats for other tasks based on this format, as shown in \textbf{Others}.}
\label{tab:mono_input}
\vspace{-1em}
\end{table*}

\begin{table*}
\centering
\scriptsize
\resizebox{0.95\linewidth}{!}{
\setlength{\tabcolsep}{4pt}
\begin{tabular}{lll}
\toprule
\bf Task& \bf Dataset & \multicolumn{1}{c}{\bf Query Instruction} \\
\midrule

\multirow{15}{*}{T$\rightarrow$T}
&ArguAna &  Given a claim, find documents that refute the claim.  \\ \cmidrule(lr){2-3}
&Climate-FEVER & Given a claim about climate change, retrieve documents that support orrefute the claim. \\ \cmidrule(lr){2-3}
&CQADupStack &  Given a question, retrieve detailed question descriptions from Stackexchange  \\ 
&&that are duplicates to the given question. \\ \cmidrule(lr){2-3}
&DBPedia & Given a query, retrieve relevant entity descriptions from DBPedia.  \\ \cmidrule(lr){2-3}
&FiQA2018 & Given a financial question, retrieve user replies that best answer the question.  \\ \cmidrule(lr){2-3}
&HotpotQA & Given a multi-hop question, retrieve documents that can help answer the question.   \\ \cmidrule(lr){2-3}
&MSMARCO & Given a web search query, retrieve relevant passages that answer the query. \\ \cmidrule(lr){2-3}
&NFCorpus &  Given a question, retrieve relevant documents that best answer the question. \\ \cmidrule(lr){2-3}
&NQ &  Given a question, retrieve Wikipedia passages that answer the question.  \\ \cmidrule(lr){2-3}
&Quora &  Given a question, retrieve questions that are semantically equivalentto the given question. \\ \cmidrule(lr){2-3}
&SCIDOCS & Given a scientific paper title, retrieve paper abstracts that are cited bythe given paper.  \\ \cmidrule(lr){2-3}
&SciFact & Given a scientific claim, retrieve documents that support or refute theclaim. \\ \cmidrule(lr){2-3}
&Touche2020 & Given a question, retrieve detailed and persuasive arguments that answer the question.  \\ \cmidrule(lr){2-3}
&TRECCOVID &  Given a query on COVID-19, retrieve documents that answer the query. \\ 
\midrule

I$\rightarrow$I 
&Nights & Find a day-to-day image that looks similar to the provided image. \\
\midrule

\multirow{4}{*}{T$\rightarrow$I}
&VisualNews & Identify the news-related image in line with the described event.  \\ \cmidrule(lr){2-3}
&Fashion200k & Based on the following fashion description, retrieve the best matching image.  \\ \cmidrule(lr){2-3}
&Memotion & \multirow{2}{*}{Retrieve the meme based on the given caption.}  \\ 
&HatefulMemes &   \\ 
\midrule

\multirow{5}{*}{T$\rightarrow$VD}
&TAT-DQA & \multirow{5}{*}{Find a screenshot that relevant to the user's question.} \\ 
&ArxivQA \\
&DocVQA  \\ 
&\textit{MITTI} \\ 
&\textit{WER}\\
\midrule

\multirow{5}{*}{I$\rightarrow$T}
&VisualNews & Find a caption for the news in the given photo.  \\ \cmidrule(lr){2-3}
&Fashion200k &  Find a product description for the fashion item in the image.\\ \cmidrule(lr){2-3}
&GLDv2 &  Retrieve the name of the landmark based on the given image.\\ \cmidrule(lr){2-3}
&Memotion &  \multirow{2}{*}{Retrieve the caption based on the given meme.}  \\ 
&HatefulMemes &  \\
\midrule 

\multirow{2}{*}{T$\rightarrow$IT}
&WebQA & Find a Wikipedia image that answers this question.  \\ \cmidrule(lr){2-3}
&EDIS & Find a news image that matches the provided caption.  \\
\midrule

\multirow{4}{*}{IT$\rightarrow$T}
&OVEN & Retrieve a Wikipedia paragraph that provides an answer to the given query about the image.   \\ \cmidrule(lr){2-3}
&INFOSEEK &  Find a paragraph from Wikipedia that answers my question about this image.  \\ \cmidrule(lr){2-3}
&OKVQA & Retrieve documents that provide an answer to the question alongside the image.  \\ \cmidrule(lr){2-3}
&VizWiz & Retrieve the correct answer for a question about an image. \\ 
\midrule

\multirow{2}{*}{IT$\rightarrow$I}
&FashionIQ & Find a fashion image that aligns with the reference image and style note.  \\ \cmidrule(lr){2-3}
&CIRR &  Retrieve a day-to-day image that aligns with the modification instructions of the provided image. \\ 
\midrule

\multirow{3}{*}{IT$\rightarrow$IT}
&OVEN & Retrieve a Wikipedia image-description pair that provides evidence for the question of this image. \\ \cmidrule(lr){2-3}
&INFOSEEK & Find an image and subject description from Wikipedia that answers my question about this image. \\ \cmidrule(lr){2-3}
&E-VQA & Obtain illustrated documents that correspond to the inquiry alongside the provided image.   \\

\bottomrule
\end{tabular}}
\caption{The instructions for different tasks. We set the instructions for the GME-2B and GMR series models on each task as shown in the Table. \textit{WER} means World Economic Reports, and \textit{MITTI} means MIT Tissue Interaction.}
\label{tab:task_instruction}
\end{table*}

\newpage
\section{Main Result}
\subsection{Detailed Results}
\label{ap:main_result}
We evaluate all models described in \S\ref{sec:exp} on our benchmark. The evaluation metrics and the detailed results for each dataset are reported in Table \ref{tab:main_table_detail}.
\begin{table*}[htbp]
\centering

% 'l' for left-aligned text, 'c' for centered numbers. Total 10 columns.
\resizebox{0.95\textwidth}{!}{\begin{tabular}{@{}llcccccc@{}}
\toprule
% Header row 1: Defines the main column groups. 'Model' spans 8 columns.
\multirow{2}{*}{\textbf{Class}} & \multirow{2}{*}{\textbf{Dataset}} & \multicolumn{6}{c}{\textbf{Model}} \\
% A rule only under the 'Model' group (columns 3 to 10).
\cmidrule(lr){3-8}
% Header row 2: Specifies the individual model names.
& & \textbf{GME-2B} & \textit{Qwen3} & \textit{MonoQwen} & \textit{Jina-m0} & \textbf{GMR-3B} & \textbf{GMR-7B}  \\
\midrule
% The 'T→T (14)' cell spans 14 rows of data.
\multirow{14}{*}{T$\rightarrow$T (14)} 
& ArguAna$^{\dag}$ & 47.11 & 86.00 & 50.93 & 56.07 & 80.42 & 84.49 \\
& SCIDOCS$^{\dag}$ & 22.65 & 26.42 & 18.31 & 22.12 & 25.49 & 28.77  \\
& TRECCOVID$^{\dag}$ & 79.11 & 87.83 & 79.84 & 85.36 & 87.23 & 85.56  \\
& Quora$^{\dag}$ & 87.35 & 88.16 & 82.71 & 87.98 & 89.51 & 89.91  \\
& SciFact$^{\dag}$ & 66.53 & 79.83 & 74.94 & 79.18 & 77.52 & 79.70 \\
& NFCorpus$^{\dag}$ & 36.90 & 41.88 & 38.29 & 40.99 & 40.51 & 40.81  \\
& Climate-FEVER$^{\dag}$ & 32.15 & 49.08 & 19.78 & 34.33 & 50.14 & 50.26  \\
& FiQA2018$^{\dag}$ & 46.35 & 56.25 & 44.11 & 50.72 & 54.79 & 59.64  \\
& HotpotQA$^{\dag}$ & 70.45 & 82.66 & 71.64 & 80.49 & 82.86 & 83.84  \\
& DBPedia$^{\dag}$ & 43.17 & 52.69 & 41.75 & 49.60 & 52.99 & 53.96  \\
& Touche2020$^{\dag}$ & 33.18 & 43.00 & 36.71 & 38.40 & 32.17 & 37.26  \\
& NQ$^{\dag}$ & 51.22 & 63.33 & 49.08 & 62.06 & 62.49 & 66.48  \\
& MSMARCO$^{\dag}$ & 40.79 & 44.57 & 35.57 & 43.09 & 45.90 & 47.60  \\
& CQADupStack$^{\dag}$ & 37.25 & 45.18 & 40.83 & 44.66 & 47.10 & 46.81  \\
\midrule
\multirow{1}{*}{I$\rightarrow$I (1)} 
& Nights$\star$ & 30.75 & - & 12.59 & 27.50 & 29.76 & 32.83 \\
\midrule
\multirow{4}{*}{T$\rightarrow$I (4)} 
& Fashion200k$^{*}$ & 25.77 & - & 29.14 & 29.38 & 25.01 & 27.57  \\
& HatefulMemes$^{\dag}$ & 52.09 & - & 74.93 & 76.57 & 75.07 & 75.19  \\
& Memotion$^{\dag}$ & 77.41 & - & 93.47 & 93.40 & 93.17 & 93.52  \\
& VisualNews$\star$ & 38.55 & - & 37.39 & 38.48 & 42.16 & 48.44  \\
\midrule
\multirow{5}{*}{T$\rightarrow$VD (5)} 
& TAT-DQA$\dagger$ & 71.23 & - & 79.99 & 82.05 & 83.23 & 84.00  \\
& DocVQA$\dagger$ & 56.44 & - & 57.51 & 61.69 & 61.48 & 62.87  \\
& ArxivQA$\dagger$ & 84.21 & - & 87.61 & 89.38 & 88.99 & 90.99  \\
& \textit{WER}$\dagger$ & 58.78 & - & 63.00 & 63.47 & 62.13 & 61.00  \\
& \textit{MITTI}$\dagger$ & 61.29 & - & 68.32 & 69.07 & 66.06 & 65.82  \\
\midrule
\multirow{5}{*}{I$\rightarrow$T (5)} 
& Fashion200k$^{*}$ & 27.67 & - & 7.55 & 17.14 & 26.22 & 29.80  \\
& HatefulMemes$^{\dag}$ & 57.85 & - & 32.27 & 80.90 & 81.21 & 81.23  \\
& Memotion$^{\dag}$ & 80.01 & - & 44.74 & 94.84 & 96.08 & 96.68  \\
& GLDv2$^{\dag}$ & 59.28 & - & 5.72 & 59.21 & 68.68 & 76.74  \\
& VisualNews$\star$ & 38.28 & - & 7.83 & 25.05 & 43.12 & 48.60  \\
\midrule
\multirow{2}{*}{T$\rightarrow$IT (2)} 
& WebQA$\star$ & 83.03 & - & 87.30 & 87.14 & 86.98 & 87.46  \\
& EDIS$\star$ & 71.00 & - & 65.63 & 62.76 & 76.95 & 81.64  \\
\midrule
\multirow{4}{*}{IT$\rightarrow$T (4)} 
& OKVQA$^{*}$ & 29.71 & - & 20.13 & 30.34 & 37.71 & 40.09  \\
& VizWiz$^{\dag}$ & 29.56 & - & 5.11 & 20.36 & 35.96 & 41.29  \\
& INFOSEEK$\star$ & 39.77 & - & 23.97 & 36.84 & 59.17 & 63.01  \\
& OVEN$\star$ & 60.46 & - & 8.18 & 23.74 & 62.41 & 68.78  \\
\midrule
\multirow{2}{*}{IT$\rightarrow$I (2)} 
& FashionIQ$^{*}$ & 26.57 & - & 21.41 & 25.97 & 30.70 & 33.32  \\
& CIRR$\star$ & 46.83 & - & 42.09 & 49.33 & 57.24 & 61.46  \\
\midrule
\multirow{3}{*}{IT$\rightarrow$IT (3)} 
& INFOSEEK$\star$ & 44.61 & - & 35.39 & 53.28 & 73.89 & 76.31  \\
& E-VQA$\star$ & 79.11 & - & 55.81 & 61.21 & 84.66 & 86.08  \\
& OVEN$\star$ & 76.96 & - & 16.28 & 40.12 & 78.68 & 84.17  \\
\bottomrule
\end{tabular}}
\caption{Detailed scores of each model on various datasets on \emph{MRB}. \textit{Qwen3} stands for Qwen3-Reranker, \textit{MonoQwen} stands for MonoQwen2-VL-v0.1, \textit{Jina-m0} stands for Jina-Reranker-m0. \textit{WER} means World Economic Reports, and \textit{MITTI} means MIT Tissue Interaction. For the datasets denoted with $\star$, we report the Recall@5 metric. Correspondingly, the Recall@10 metric is adopted for the datasets marked with $^{*}$. Furthermore, the NDCG@5 score is utilized for the $\dagger$-annotated datasets, while the NDCG@10 score is reported for those designated with $^{\dag}$.}
% For the datasets with $\star$, we report Recall@5 scores. For the datasets with $^{*}$, we report Recall@10 scores. For the datasets with $\dagger$, we report NDCG@5 scores. For the datasets with $^{\dag}$, we report NDCG@10 scores.
\label{tab:main_table_detail}
\end{table*}

\subsection{The influence of the number of negative}
\label{ap:main_result:neg_num}
\begin{wrapfigure}{r}{0.4\textwidth}  
\vspace{-1.5em}  
\includegraphics[width=0.4\textwidth]{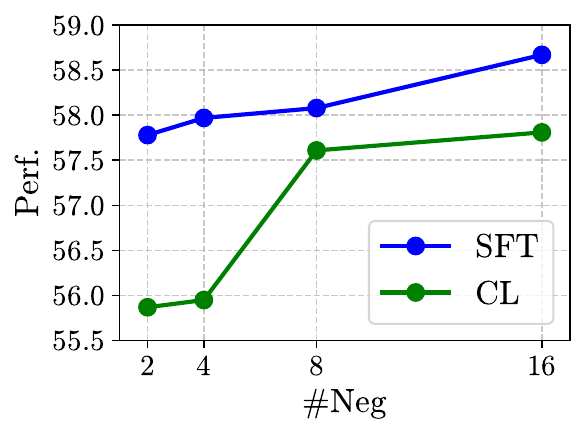}
\vspace{-1.75em}  
\caption{Average performance of the number of negatives per sample.}
\label{fig:negNum}
\end{wrapfigure}
In \S \ref{sect:main_result}, we examine the effect of incorporating negatives in supervised fine-tuning (SFT) and observed that, within the limits of available computational resources, increasing the number of negative examples consistently improved model performance. The best performance was achieved when the number of negative examples reached 16. For comparison, we further conduct experiments on the role of negatives in contrastive learning. As shown in \Figref{fig:negNum}, the results indicate that, similar to SFT, a larger number of negative examples leads to better performance. Nevertheless, the overall performance of contrastive learning remains lower than that of supervised fine-tuning.

\begin{wraptable}{r}{0.35\textwidth}
\vspace{-1em}
\begin{tabular}{cccc}
\toprule
& \textit{-F} & \textit{-NF} & $\Delta_f$ \\
\midrule
SFT & 57.97 & 57.94 & {\textcolor{red}{$\blacktriangledown$}~0.03}  \\
\midrule
CL  & 55.95 & 57.20 & {\textcolor{green}{$\blacktriangle$}~1.25}  \\
% \midrule
% $\Delta$ & 2.02 & 0.74 \\
\bottomrule
\end{tabular}
\caption{Impact of frozen of the LM head on performance. \textit{-F} denotes frozen, while \textit{-NF} denotes not frozen.}
\vspace{-1em}
\label{tab:nofreeze}
\end{wraptable}

\subsection{The influence of the frozen of LM head}
\label{ap:main_result_nofreeze}
In \S \ref{sec:analysis}, we observe that SFT can exploit semantic signals from pre-trained token embeddings, whereas CL must learn the score-projection matrix from scratch. To rule out the potential influence of freezing the language modeling (LM) head parameters, we conduct an ablation study on LM head parameter freezing, with the results presented in Table \ref{tab:nofreeze}. The findings show that freezing or unfreezing the LM head has no effect on SFT. In contrast, CL achieves better performance when the LM head parameters are not frozen. These results suggest that SFT effectively leverages the semantic information embedded in pre-trained token of LLM, while CL requires relearning the score-projection matrix.

\section{Limitation}
In this work, we introduce MRB, a benchmark designed for training and evaluating multimodal reranking tasks. To address this challenge, we investigate strategies for adopting Multimodal Large Language Models (MLLMs) into general-purpose multimodal reranking models, and propose GMR, a reranking model capable of handling candidates across different modalities. Despite these contributions, our work has the following limitations:

$\bullet$ Single-language constraint.
Although the backbone model, Qwen2.5-VL-Instruction, supports multiple languages, we trained and evaluated GMR exclusively in English. Consequently, the performance of GMR in other languages remains unexplored.

$\bullet$ Single-image constraint for queries and documents.
For reasons of training efficiency and limited availability of relevant data, both queries and candidates in MRB are restricted to a single image for each query and document. As a result, the benchmark cannot assess performance on interleaved inputs that involve multiple images and texts.

\end{document}